\documentclass{article} 
\usepackage{iclr2024_conference,times}

\usepackage{amsmath,amsfonts,bm}

\def\eqref#1{equation~\ref{#1}}

\def\1{\bm{1}}

\DeclareMathAlphabet{\mathsfit}{\encodingdefault}{\sfdefault}{m}{sl}
\SetMathAlphabet{\mathsfit}{bold}{\encodingdefault}{\sfdefault}{bx}{n}

\usepackage{url}
\usepackage{mathabx}
\usepackage{amsmath,amssymb,amsthm}
\usepackage[acronym]{glossaries}
\usepackage{letltxmacro}
\usepackage{nicefrac}
\usepackage{booktabs}
\usepackage{multirow}

\usepackage[capitalize,nameinlink]{cleveref}

\usepackage{graphicx}
\usepackage{subcaption}

\usepackage{fontawesome}

\usepackage{xcolor}
\definecolor{darkgreen}{RGB}{0,51,0}
\definecolor{dtured}{RGB}{153,0,0}

\usepackage[title]{appendix}

\usepackage{tikz}
\usepackage{pgfplots}
\pgfplotsset{compat=newest}

\usepackage{algorithm}
\usepackage{algpseudocode}
\algrenewcommand\algorithmicrequire{\textbf{Input:}}
\algrenewcommand\algorithmicensure{\textbf{Output:}}

\newcommand{\eqrefp}[1]{(\ref{#1})}

\newtheorem{theorem}{Proposition}
\def\thm{Proposition}

\newacronym{iot}{IoT}{internet of things}
\newacronym{miqp}{MIQP}{mixed integer quadratic programming}
\newacronym{ot}{OT}{optimal transport}
\newacronym{fim}{FIM}{Fisher information matrix}
\newacronym{mmd}{MMD}{maximum mean discrepancies}
\newacronym{fl}{FL}{federated learning}
\newacronym{wrp}{WRP}{Wasserstein Regression Pruning}
\newacronym{sgd}{SGD}{stochastic gradient descent}
\newacronym{iht}{IHT}{iterative hard thresholding}
\newacronym{ewr}{EWR}{entropic Wasserstein regression}
\newacronym{lr}{LR}{linear regression}

\makeatletter
\let\oldmakefirstuc\makefirstuc
\renewcommand*{\makefirstuc}[1]{%
  \def\gls@add@space{}%
  \mfu@capitalisewords#1 \@nil\mfu@endcap
}
\def\mfu@capitalisewords#1 #2\mfu@endcap{%
  \def\mfu@cap@first{#1}%
  \def\mfu@cap@second{#2}%
  \gls@add@space
  \oldmakefirstuc{#1}%
  \def\gls@add@space{ }%
  \ifx\mfu@cap@second\@nnil
    \let\next@mfu@cap\mfu@noop
  \else
    \let\next@mfu@cap\mfu@capitalisewords
  \fi
  \next@mfu@cap#2\mfu@endcap
}
\makeatother

\title{\texttt{SWAP}: \texttt{S}parse Entropic \texttt{WA}sserstein Regression for Robust Network \texttt{P}runing}

\author{Lei You\thanks{Correspondence to both \texttt{leiyo@dtu.dk} and \texttt{hvc@ece.au.dk}. Lei You is supported by Thomas B. Thriges Fond 5041-2402. Hei Victor Cheng is supported by the Aarhus Universitets Forskningsfond under Project AUFF 39001.}\\
Department of Engineering Technology\\
Technical University of Denmark\\
Ballerup, DK-2750, Denmark \\
\texttt{leiyo@dtu.dk} \\
\And
Hei Victor Cheng$^*$\\
Department of Electrical and Computer Engineering \\
Aarhus Universrity \\
Aarhus, DK-8200, Denmark \\
\texttt{hvc@ece.au.dk} 
}

\newcommand\norm[1]{\left\lVert#1\right\rVert}

\renewcommand{\vec}[1]{\mathbf{#1}}
\renewcommand{\d}[1]{\ensuremath{\operatorname{d}\!{#1}}}

\def\L{\mathcal{L}}

\def\P{\mathcal{P}}
\def\RR{\mathbb{R}}
\def\EE{\mathbb{E}}
\def\VV{\mathbb{V}}
\def\UU{\mathbb{U}}

\def\B{\mathcal{B}}
\def\N{\mathcal{N}}
\def\bw{\bar{w}}

\def\ot{W_2^2}
\def\nel{{\nabla\bm{\ell}}}
\DeclareMathOperator{\diag}{diag}

\iclrfinalcopy 
\begin{document}
\usetikzlibrary{pgfplots.statistics, pgfplots.colorbrewer} 
\usepgfplotslibrary{colorbrewer}

\maketitle

\begin{abstract}
This study addresses the challenge of inaccurate gradients in computing the empirical Fisher Information Matrix during neural network pruning. We introduce \texttt{SWAP}, a formulation of Entropic Wasserstein regression (EWR) for pruning, capitalizing on the geometric properties of the optimal transport problem. The ``swap'' of the commonly used linear regression with the EWR in optimization is analytically demonstrated to offer noise mitigation effects by incorporating neighborhood interpolation across data points with only marginal additional computational cost. The unique strength of \texttt{SWAP} is its intrinsic ability to balance noise reduction and covariance information preservation effectively.
Extensive experiments performed on various networks and datasets show comparable performance of \texttt{SWAP} with state-of-the-art (SoTA) network pruning algorithms. Our proposed method outperforms the SoTA when the network size or the target sparsity is large, the gain is even larger with the existence of noisy gradients,  possibly from noisy data, analog memory, or adversarial attacks. Notably, our proposed method achieves a gain of 6\% improvement in accuracy and 8\% improvement in testing loss for MobileNetV1 with less than one-fourth of the network parameters remaining.  
\end{abstract}

\section{Introduction}

The advent of deep learning has revolutionized various domains of artificial intelligence, with neural networks showing remarkable performance across an array of applications. Nonetheless, the increase in model complexity has led to escalating computational demands and substantial memory requirements. This poses significant challenges for deploying these models in resource-constrained environments such as mobile or \gls{iot} devices. Therefore, the concept of neural network pruning emerges as a critical solution. It aims to optimize the network by removing less important parameters, which reduces computational overhead while maintaining the performance of the original model.

In the realm of state-of-the-art deep learning, the models often exhibit substantial size and complexity, with up to trillions of parameters, as exemplified by models such as GPT-4. The immense computational demand, energy inefficiency, and the challenges with model interpretability associated with these models highlight the need for innovative and efficient optimization techniques. These techniques should ideally minimize the model size while improving their robustness and interpretability. Considering the limitations of previous work, especially those arising from the influence of noisy data and noisy gradients, the paper proposes a promising pathway for robust pruning. 

Below, we inspect the network pruning problem from an optimization perspective, with a concise introduction of the most relevant existing works. Then a sketch of our approach is given.

\textbf{Related Work on Pruning as Optimization.} Denote by $\vec{\bw}\in\RR^p$ a trained model and $\L(\vec{w})$ the loss function given arbitrary model $\vec{w}$.
The loss function can be locally approximated around $\vec{\bw}$ with Taylor Expansion as shown in \eqrefp{eq:taylor}.
\begin{equation}
\L(\vec{w}) = \L(\vec{\bw}) + \nabla\L(\vec{\bw})^{\top}(\vec{w}-\vec{\bw}) +\frac{1}{2}(\vec{w}-\vec{\bw})^{\top}\nabla^2\L(\vec{\bw})(\vec{w}-\vec{\bw}) + O(\norm{\vec{w}-\vec{\bw}}^3)
\label{eq:taylor}
\end{equation}
Consider a neural network with a loss function $\L(\vec{w})=\frac{1}{N}\sum_{i=1}^N {\bm{\ell}}_i(\vec{w})$, where $\bm{\ell}_i(\vec{w})\in\RR^{p}$ is the loss incurred at data point $i$ ($i=1,\ldots,N$). The goal of network pruning is to find a set of $\vec{w}$ such that there are $k$ ($k<p$) elements of $\vec{w}$ being zero while keeping the newly obtained model $\vec{w}$'s performance as good as possible to the original one $\vec{\bw}$. Mathematically, we want to find some $\vec{w}\in\RR^p$ that satisfies both $\L(\vec{w})\approx\L(\vec{\bw})$ and $\norm{\vec{w}}_0\leq k$, with $k<p$.

This line of research can be dated back to \citep{lecun1989optimal}, where the approximation in equation \eqrefp{eq:taylor} is adopted. Under the assumption that gradient $\nabla \L(\vec{\bw})\approx 0$ when the network is trained, the network weights are pruned one-by-one in decreasing order based on the value of $(\vec{w}-\vec{\bw})^\top\vec{H}(\vec{w}-\vec{\bw})$.
In their approach, the $\vec{H}$ is approximated as a diagonal matrix; this is later extended in
\citep{hassibi1992second} to include the whole Hessian matrix, and the authors also proposed using the \gls{fim} as an approximation to the Hessian. Later, \citep{singh2020woodfisher} proposed to reduce the computation complexity by using block diagonal Hessian, and \gls{fim} is approximated using a small subset of the training data.

These approaches all use equation \eqrefp{eq:taylor} to prune the network in a one-by-one manner, namely the weight with the least importance is set to zero according to the different approximations of equation \eqrefp{eq:taylor}. In this way, the potential interactions of pruning multiple weights are ignored. To explore this, the network pruning problem is formulated as a \gls{miqp} in \citep{yu2022combinatorial}. Namely, an objective function 
\begin{equation}
f(\vec{w})=(\vec{w}-\vec{\bw})^\top\vec{H}(\vec{w}-\vec{\bw}) + \lambda\norm{\vec{w}-\vec{\bw}}^2\text{ ($\lambda \geq 0$)}
\label{eq:miqp}
\end{equation}
is minimized, and Hessian is approximated as $\vec{H}\approx \nabla^2\L(\vec{\bw})$, subject to the sparsity constraint $\norm{\vec{w}}_0\leq k$, where $\lambda$ is a regularization parameter. Although this approach shows significant improvements, it suffers from scalability issues as a full Hessian matrix is required.

\textbf{Sparse \Gls{lr} Formulation}. To reduce the computational complexity, the Hessian matrix can be approximated by the empirical \gls{fim}, using $n$ samples as in \citep{chen2022network,benbaki2023fast}. Denote $\vec{G}=[\nabla{\bm{\ell}}_1,\ldots,{\nel}_n]^\top \in \RR^{n\times p}$, where $\nabla \bm{\ell}_i=\nabla \bm{\ell}_i(\vec{\bw})$. For simplicity, $\nabla \bm{\ell}_i$ is used in this document to represent the derivative of the data point $i$'s loss at $\vec{\bw}$ consistently in this paper unless specified otherwise. The Hessian is approximated through the expression $\vec{H}\approx (1/n)\sum_{i=1}^n\nel_i \nel_i^{\top}=(1/n)\vec{G}^\top \vec{G}$, which is the so-called \gls{fim}. Denote $x_i=\nel_i^\top\vec{w}$ and $y_i=\nel_i^\top\vec{\bw}$, \eqrefp{eq:miqp} is formulated to the sparse \gls{lr} problem shown in \eqrefp{eq:Euclidean_formulation} below.
\begin{equation}
                    \min_{\vec{w}} \bar{Q}(\vec{w}) = \sum_{i=1}^n \norm{x_i(\vec{w})-y_i}^2  + n\lambda \norm{\vec{w} - \vec{\bw}}^2, 
\text{ s.t. }\|\vec{w}\|_{0} \leq k   
\label{eq:Euclidean_formulation}
\end{equation}
This formulation has a computational advantage, as empirical FIM needs not to be computed explicitly. It is shown that the formulation scales \textit{to large neural network pruning} \citep{chen2022network}.

\textbf{Motivation of Combating Against Noise}. 
In practice, it is not always easy to obtain the correct gradients for pruning large neural networks. There can be noise contained in the data samples, and the gradients can also be corrupted due to various reasons, e.g., distributed or federated learning \citep{turan2022robust}, or adversarial attacks such as data poisoning \citep{steinhardt2017certified}. 

As pointed out by \citep{mahsereci2017early, siems2021dynamic}, conditioning on the underlying true gradient $\nabla \L(\vec{\bw}) = 0$, there are mini-batch gradients which are not informative anymore as it can be fully explained by sample noise and the
vanishing gradients. These gradients would not contribute to the covariance information of the empirical \gls{fim} but serve as outliers in Hessian approximation. 

In the scenarios of \gls{fl}, gradients
computed by different clients are skewed and consequently, local models move
away from globally optimal models \citep{huang2022fedtiny}, imposing challenges for constructing informative \gls{fim}. Besides, noise can be added to the gradient for privacy concerns in communications \citep{li2020federated}. Additionally, the clients usually have inevitable noisy samples and labels, making models suffer from a significant performance drop \citep{tuor2021overcoming}. Additionally, over-the-air communications itself suffer from unavoidable noises \citep{ang2020robust,yang2020federated}. These lead to concerns for network pruning with noisy gradients.
Finally, analog memory recently gained attention for deep learning model deployment \citep{garg2022dynamic}. When neural network parameters and data are stored in these analog devices, they are susceptible to device-related noise, affecting the performance of network compression \citep{isik2023neural}.

\textbf{Approach Sketch}. We revisit the \gls{miqp} network pruning optimization from a perspective of \gls{ewr}, which leverages Wasserstein distance to model the dissimilarity between two distributions. In our context, it measures 
the dissimilarity of distributions relevant to model parameters and gradient magnitudes before and after pruning. Namely, $\nel$ is a $p$ dimensional distribution, capturing geometric properties of the loss at $\vec{\bw}$ before pruning. Both $\vec{\bw}$ and $\vec{w}$ perform projections for $\nel$ to a 1-D distribution respectively as $\nel^\top\vec{\bw}$ and $\nel^\top\vec{w}$. Computing the distance between $\nel^\top\vec{\bw}$ and $\nel^\top\vec{w}$ falls into the framework of sliced probability divergence \citep{nadjahi2020statistical}. Under this framework, pruning optimization essentially fine-tunes $\vec{w}$ and selectively reserves its elements such that the divergence is minimized subject to the sparsity constraint. 

Our approach's effectiveness in combating noisy gradients is established both analytically and numerically. We demonstrate that \textit{pruning through the Wasserstein regression implicitly enacts gradient averaging using Neighborhood Interpolation}. This entails a nuanced balance between capturing gradient covariance and diminishing gradient noise. Notably, the sparse \gls{lr} formulation is merely a specific instance of ours. Yet, our proposed algorithm doesn't demand a markedly higher computational expense. This modest additional effort bestows upon us enhanced robustness.

\section{Problem Setup and Formulation}
\label{sec:formulation}

We first introduce the \gls{ot} problem in Kantorovich formulation with entropic regularization, which measures the distance between two distributions, defined in \eqrefp{eq:Wasserstein} below. The Wasserstein regression formulation as a generalization of~the \gls{lr} formulation is then proposed.

\textbf{The Kantorovich Problem}. Denote $\P_2$ the set of probability measures with finite second moments. Let $\mu$, $\nu\in\P_2$ and let $\Pi(\mu, \nu)$ denote the set of probability measures in $\P_2$ with marginal distributions equal to $\mu$ and $\nu$. The 2-Wasserstein distance is defined as 
\begin{equation}
\ot(\mu, \nu) = \inf_{\pi\in\Pi(\mu, \nu)} \int_{\RR^{d\times d}} \norm{x-y}^2 \d \pi(x,y) + \varepsilon\int_{\RR^{d\times d}}\log\left(\frac{\d\pi}{\d\mu\d\nu}\right)\d\pi. 
\label{eq:Wasserstein}%
\end{equation}
This is also referred to as the entropic \gls{ot} problem, where the first term is the transportation cost between the two measures and the second term is the entropic regularization with multiplier $\varepsilon$.

\textbf{Sparse \gls{ewr} Formulation}. The pruning problem formulation is defined in \eqrefp{eq:wasserstein_formulation} below.
\begin{subequations}
\begin{align}
         \min_{\vec{w}}   & \quad Q(\vec{w}) = \ot(x(\vec{w}),y) + \lambda\norm{\vec{w} - \vec{\bw}}^2  \\
\text{ s.t. } & \quad \norm{\vec{w}}_{0} \leq k
\end{align}
\label{eq:wasserstein_formulation}%
\end{subequations}
The term $\ot(x(\vec{w}),y)$ is a Wasserstein distance between the two one-dimensional distributions $x$ and $y$ (or a sliced Wasserstein distance for $\nabla\ell$ with two one-dimensional projections). The optimization is to alter $\vec{w}$ such that the distance between the two distributions is minimized.

Let $x$ and $y$ follow the empirical distributions $\{x_i\}_{i=1}^n$ and $\{y_i\}_{i=1}^n$. Denote by $\mu_i$ and $\nu_i$ the mass of the data points $x_i$ and $y_i$, respectively. We use $\vec{\Pi}$ to refer to a matrix representing the transportation probability between $x$ and $y$, and $\Pi$ the set of all such matrices, i.e. $\Pi=\{\vec{\Pi} | \sum_{i=1}^n \pi_{ij} = \mu_j~\forall j \text{ and } \sum_{j=1}^n \pi_{ij}=\nu_i~\forall i\}$, where $\mu_i$ and $\nu_j$ are marginal distributions. Then~\eqrefp{eq:wasserstein_formulation} reads:
\begin{subequations}
\begin{align}
\min_{\vec{w}}  &~   Q(\vec{w}) = \!\!\inf_{\vec{\Pi}\in\Pi}\left\{\sum_{i=1}^n\sum_{j=1}^n\norm{x_i(\vec{w})-y_i}^2\pi_{ij} + \varepsilon\sum_{i=1}^n\sum_{j=1}^n\log\left( \frac{\pi_{ij}}{\mu_i \nu_j} \right) \pi_{ij} \right\} + \lambda\norm{\vec{w} - \vec{\bw}}^2 \\
\text{s.t. } &~ \norm{\vec{w}}_{0} \leq k   
\end{align}
\label{eq:wasserstein_formulation_empirical}%
\end{subequations}

\textbf{\Gls{lr} as a Special Case}. Let $\varepsilon=0$. Once we set $\vec{\Pi}$ to be a diagonal matrix with constant value $1/n$, i.e. $\diag(1/n)$, the mass transportation happens only between data point pairs $(x_i, y_i)$  for $i=1,\ldots,n$. Therefore we have
\begin{equation}
Q_{\vec{\Pi}=\diag(1/n)}(\vec{w}) = \frac{1}{n} \sum_{i=1}^n \norm{x_i(\vec{w})-y_i}^2 + \lambda\norm{\vec{w} - \vec{\bw}}^2 = \frac{1}{n} \bar{Q}(\vec{w}),
\label{eq:special_case}%
\end{equation}
i.e. the formulation~\eqrefp{eq:wasserstein_formulation_empirical} in this case degrades to the \gls{lr} formulation in~\eqrefp{eq:Euclidean_formulation}.

\section{Theoretical Aspects}
\label{sec:theoretical}

This section reveals some good theoretical properties of the Sparse \gls{ewr} formulation for network pruning. We start with Proposition~\ref{thm:convex_equality} below (proof in Appendix~\ref{sec:proof_convex_equality}) that states a geometry property of \gls{ot} with squared Euclidean distance cost. Additionally, we demonstrate the Neighborhood Interpolation mechanism that happens implicitly in solving the \gls{ewr}. Moreover, we show that such a mechanism strikes a balance in capturing gradient covariance and reducing gradient noise, with a brief discussion on the advantage of using entropic regularization in terms of sample complexity.

\begin{theorem}[Convex Hull Distance Equality]
Consider a set $S$ and its convex hull $\textit{Conv}(S)$ in a Euclidean space, and an arbitrary point ${x}$ in the space. For any probability measure $\hat{\nu}$ on $S$, we can find a point ${y}'$ in $\textit{Conv}(S)$ as ${y}' = \int {y} \, \mathrm{d}\nu({y})$ such that $\|{x} - {y}'\|^2 = \int \|{x} - {y}\|^2 \, \mathrm{d}\hat{\nu}({y})$, where $\nu$ is a measure on $\textit{Conv}(S)$.
\label{thm:convex_equality}
\end{theorem}

\textbf{Neighborhood Interpolation}. In formulation~\eqrefp{eq:wasserstein_formulation}, let $W_x$ be the first term of $W_2^2$ for an arbitrary given $x$, i.e., $W_x = \int\norm{x(\vec{w})-y}^2 \d \pi(\cdot|x)(y)$, where $\pi(\cdot|x)(y)$ is a conditional measure given $x$. 
Now, divide the Euclidean space $\RR^d$ by subspaces $S^y_1$, $S^y_2$\ldots, $S^y_n$ for $y$. For any conditional measure $\pi(\cdot|x)(y)$ defined on any $S^y_i$ ($i=1,\ldots,n$), there exists a measure $\nu(y)$ defined on $\textit{Conv}(S_i^y)$ such that the weighted distance from $S^y_i$ to $x$ equals the distance from $x$ to a point $y'$ in $\textit{Conv}(S^y_i)$. Hence
\begin{align*}
    W_x & = \int_{S^y_1\cup S^{y}_2\cup \cdots \cup S^{y}_n}\left\|{x}-y \right\|^2 \d \pi(\cdot|x)(y) \\
                 & = \frac{1}{n}\sum_{i=1}^n\left\|{x}-y_i' \right\|^2 \text{ s.t. }y_i'=\int_{\textit{Conv}(S^y_i)} {y}\,\d{\nu(y)}, \nu\in \mathbb{V}_i(x),~i=1,\ldots n.  
\end{align*}%
where $\VV_i(x)$ is the set of measures $\nu$ that make the equality holds and $\VV_i(x)\neq \varnothing$ by \thm~\ref{thm:convex_equality}.

Similarly, we define $W_y$ for any given $y$ and subspaces $S^x_1$, $S^x_2\ldots, S^x_n$, 
\begin{align*}
    W_y & = \int_{S^x_1\cup S^{x}_2\cup \cdots \cup S^{x}_n}\left\|{x}-y \right\|^2 \d \pi(\cdot|y)(x) \\
                 & = \frac{1}{n}\sum_{i=1}^n\left\|{x}'_i-y \right\|^2 \text{ s.t. }x'_i=\int_{\textit{Conv}(S^x_i)} {x}\,\d\mu{(x)},~\mu\in \mathbb{U}_i(y),~i=1,\ldots n.  
\end{align*}
where $\mu$ is a measure defined on $\textit{Conv}(S^x_i)$ and $\UU_i(y)\neq \varnothing$.

\begin{figure}[H]
\begin{minipage}{0.3\textwidth}
\centering
\scalebox{0.9}{
\begin{tikzpicture}
    \draw[thick,fill=lightgray, fill opacity=0.3] (0,0) -- (4,0) -- (2,3) -- cycle; 
    \draw[fill] (0,0) circle [radius=0.1] node[below left] {${y_1}$}; 
    \draw[fill] (2,0) circle [radius=0.1] node[below] {${y_2}$}; 
    \draw[fill] (1,1) circle [radius=0.1] node[below] {${y_3}$}; 
    \draw[fill] (4,0) circle [radius=0.1] node[below] {${y_4}$}; 
    \draw[fill] (2,3) circle [radius=0.1] node[above] {${y_5}$}; 
    \draw[fill=blue] (0.5,2) circle [radius=0.1] node[above left] {${x}$}; 
    \draw[fill=red] (1.7,1.2) circle [radius=0.1] node[below right] {${y}'$}; 
    \draw[dashed] (0.5,2) -- (1.7,1.2); 
    \draw[thick,dotted,red] (0,0) -- (1.7,1.2); 
    \draw[thick,dotted,red] (4,0) -- (1.7,1.2); 
    \draw[thick,dotted,red] (2,3) -- (1.7,1.2); 
\end{tikzpicture}
}
\end{minipage}
\hspace{0.03\textwidth}
\begin{minipage}{0.63\textwidth}
\textit{\footnotesize
We demonstrate the concept of ``Neighborhood Interpolation'' through an empirical distribution example. Define \scalebox{0.9}{$ S $} as a subset of \scalebox{0.9}{$ y $} such that for every element \scalebox{0.9}{$ y_i \in S $}, \scalebox{0.9}{$ \pi_{x,i} > 0 $}. Without loss of generality, we can denote \scalebox{0.9}{$ S = \{y_1, y_2, y_3, y_4, y_5\} $}. The area shaded in gray, denoted as \scalebox{0.9}{$ \textit{Conv}(S) $}, represents the convex hull of \scalebox{0.9}{$ S $}. \scalebox{0.9}{$ W_x $} computes a weighted summation of distances between \scalebox{0.9}{$ x $} and the points \scalebox{0.9}{$ y_1, \ldots, y_5 $}. The weights \scalebox{0.9}{$ \pi_{x,1}, \ldots, \pi_{x,5} $} are decided by \gls{ot}. A significant \scalebox{0.9}{$ \pi_{x,i} $} typically implies that \scalebox{0.9}{$ y_i $} is in proximity to \scalebox{0.9}{$ x $}, indicating a neighborhood relation. By Proposition~\ref{thm:convex_equality}, this weighted distance is analogous to the distance between \scalebox{0.9}{$ x $} and \scalebox{0.9}{$ y' $}, where \scalebox{0.9}{$ y' $} is derived from \scalebox{0.9}{$ \textit{Conv}(S) $}.
}
\end{minipage}
\label{fig:proposition 1}
\end{figure}


\textbf{Revisit the \gls{ewr} formulation}. The integral of either $W_x$ or $W_y$ respectively on $x$ or $y$ gives the first term of $W_2^2$. One can then reformulate~\eqrefp{eq:wasserstein_formulation} as \eqrefp{eq:wasserstein_formulation_reform} below.
\begin{equation}
\min_{\vec{w}:\|\vec{w}\|_{0} \leq k } Q(\vec{w}) = \frac{1}{2}\inf_{\pi}\left\{\int W_{x}(\vec{w})  \d \mu(x) + \int W_{y}(\vec{w})  \d \nu(y)\right\} + \lambda\norm{\vec{w} - \vec{\bw}}^2.
\label{eq:wasserstein_formulation_reform}
\end{equation}
Interpretation: The objective function calculates the Euclidean distance between a point \(x\) and \(n\) distinct points. These \(n\) points originate from \(n\) convex hulls, each shaped by different \(n\) subspaces within \(y\). Similarly, the function measures the distance between each point \(y\) and \(m\) unique points derived from \(m\) convex hulls, each formed by distinct \(m\) subspaces within \(x\).

We claim that the \gls{ewr} formulation is more resilient to noisy gradient than its counterpart, the \gls{lr} formulation given by~\eqrefp{eq:Euclidean_formulation}. To understand this claim better, let us reimagine the problem using empirical distributions, as indicated by \eqrefp{eq:wasserstein_formulation_empirical}. In this context, we use \(x_i\) and \(y_i\) as substitutes for \(S^x_i\) and \(S^y_i\). Moreover, the integration in both \(W_x\) and \(W_y\) is replaced with summations, offering a more insightful version of our initial \gls{ewr} formulation, shown as~\eqrefp{eq:wasserstein_formulation_empirical_2}.
\begin{equation}
\min_{\vec{w}:\norm{\vec{w}}_0\leq k} Q(\vec{w}) = \inf_{\vec{\Pi}}\left\{Q_{\vec{\Pi}}(\vec{w}) + \varepsilon\sum_{i=1}^n\sum_{j=1}^n\log\left( \frac{\pi_{ij}}{\mu_i \nu_j} \right) \pi_{ij} \right\}
\label{eq:wasserstein_formulation_empirical_2}
\end{equation}
The notation $Q_{\vec{\Pi}}(\vec{w})$ defined in \eqrefp{eq:wasserstein_formulation_empirical_3} denotes the part of the objective function given fixed $\vec{\Pi}$:
\begin{subequations}
\begin{align}
Q_{\vec{\Pi}}(\vec{w}) & = \sum_{i=1}^n\sum_{j=1}^n\norm{x_i(\vec{w})-y_j}^2\pi_{ij}   + \lambda\norm{\vec{w} - \vec{\bw}}^2 \\
 & = \frac{1}{2}\underbrace{\sum_{i=1}^n\norm{x_i(\vec{w})-y'_i}^2}_{K_{\vec{\Pi}}^{(1)}} + \frac{1}{2}\underbrace{\sum_{i=1}^n\norm{x'_i(\vec{w})-y_i}^2}_{K_{\vec{\Pi}}^{(2)}} + \lambda\norm{\vec{w} - \vec{\bw}}^2 \label{eq:wfe_final}
\end{align}
\label{eq:wasserstein_formulation_empirical_3}%
\end{subequations}
In \( Q_{\vec{\Pi}}(\vec{w}) \), for each index \( i \), points \( x'_i \) and \( y'_i \) are chosen from the convex hulls formed by points in \( x \) and \( y \), as per the guidelines of \thm~\ref{thm:convex_equality}. Now, contrasting this with the \gls{lr} model in \eqrefp{eq:Euclidean_formulation}, the objective \(\bar{Q}(\vec{w})\) aims for regression directly over the data points whereas every point from one empirical set is matched for Euclidean distance computation to a point derived from a convex combination of the other.


The infimum in \eqrefp{eq:wasserstein_formulation_empirical_2} seeks the \gls{ot} plan, \( \vec{\Pi} \), that aligns the empirical distributions \( x \) and \( y \) closely. In practical terms, for each data point \( x_i \), only a subset of \( \{y_i\}_{i=1}^n \) will transport a substantial mass, rather than the entire set. This behavior of \( \vec{\Pi} \) effectively defines \( n \) "neighborhoods" for each data point \( x_i \) within the empirical distribution of \( y \). Here, a "neighborhood" refers to a group of data points in \( y \) that are proximate to a specific \( x_i \) in the Euclidean sense.

\textbf{Neighborhood Size Control}. A critical aspect of this formulation is the entropic regularization term, which is used to modulate the size of these neighborhoods. Specifically, increasing the value of \( \varepsilon \) amplifies the impact of the entropy term. This change broadens the neighborhoods, drawing more data points into the fold of the associated convex hulls. An illustrative extreme case is when \( \varepsilon = 0 \). Here, the \gls{ot} does one-to-one matching, implying that each data point \( y_i \) primarily forms the convex hull independently. On the contrary, when $\varepsilon\rightarrow \infty$, all data points are equally weighted by $\vec{\Pi}$ and hence involved in forming the convex hull as a neighborhood.

\textbf{Capturing Covariance With Gradient Noise Reduction}. For an arbitrary $\vec{w}$, the \gls{ewr} formulation essentially strikes a balance between gradient noise reduction and covariance capturing. 
We show the analysis for $K^{(1)}_{\vec{\Pi}}$ in \eqrefp{eq:wasserstein_formulation_empirical_3}, and $K^{(2)}_{\vec{\Pi}}(\vec{w})$ follows similarly. Note that
$y'_i = \sum_{j=1}^n\nu^{(i)}_j y_j = \sum_{j=1}^{n}\nu^{(i)}_j{\nel}^\top_j\vec{\bw}$, where $\bm{\nu}^{(i)}$ are convex combination coefficients by Proposition~1. Denote ${\nel}'^\top_i = \sum_{j=1}^{n}\nu^{(i)}_j{\nel}^\top_j$, and $\vec{G}'=[\nabla\bm{\ell}'_1,\ldots,\nabla\bm{\ell}'_n]^{\top}$. The term $K^{(1)}_{\vec{\Pi}}$ expands as follows.
\begin{align}
K^{(1)}_{\vec{\Pi}} &= \sum_{i=1}^{n}({\nel}_i^\top\vec{w}-{{\nel}'}^{\top}_i\vec{\bw})^\top ({\nel}_i^\top\vec{w}-{{{\nel}'}}_i^\top \vec{\bw}) \nonumber \\
&=\sum_{i=1}^{n}(\vec{w}^\top{\nel}_i{\nel}_i^\top\vec{w}-\vec{w}^\top{\nel}_i\nel_i'{}^\top\vec{\bw}-\vec{\bw}{}^\top\nel_i'{\nel}_i{}^\top\vec{w}+\vec{\bw}{}^\top\nel_i'\nel_i'{}^\top\vec{\bw})
\label{eq:K1}
\end{align}
Examining \(K^{(1)}_{\vec{\Pi}}\) from \eqrefp{eq:K1}, we see that it effectively replaces half of \({\nel}_i\) with \({\nel}'_i\), a version obtained through weighted gradient averaging. Now let's compare the covariance between $\nabla\bm{\ell}_i$ and $\nabla\bm{\ell}'_i$. Assume that ${\nel}_i$ ($1\leq i \leq n$) are i.i.d. with the same covariance matrix $\Sigma$, then $\vec{G}'$ is with equal or less noise than $\vec{G}$. To show this, denote the covariance matrix of each ${\nel}'_i$ by
\[
\Sigma'_i = \text{Cov}\bigg[\sum_{j=1}^n \nu^{(i)}_j {\nel}_j\bigg] = \sum_{j=1}^n [\nu^{(i)}_j]^2 \Sigma.
\]
The total variance of each gradient in $\vec{G}'$ (i.e., the trace of $\Sigma'_i$) is then
\[
\text{trace}(\Sigma'_i) = \text{trace}\bigg(\sum_{j=1}^n \big[\nu^{(i)}_j\big]^2 \Sigma\bigg) = \sum_{j=1}^n \big[\nu^{(i)}_j\big]^2 \text{trace}(\Sigma) \leq \text{trace}(\Sigma).
\]
The last inequality follows from the fact that $\sum_{j=1}^n [\nu^{(i)}_j]^2 \leq 1$, which is a consequence of the Cauchy-Schwarz inequality given that the coefficients $\nu^{(i)}_j$ form a convex combination.

Originally, the covariance information of all data points is embedded in \({\nel}_i{\nel}_i^{\top}\) for \(i=1,2,\ldots,n\). An alternative representation is \({\nel}'_i\nel_i'^{\top}\), which prioritizes noise reduction, but sacrifices some covariance information. Both \({\nel}'_i{\nel}_i^{\top}\) and \({\nel}_i\nel_i'^{\top}\) highlight a trade-off. Notably, both the original covariance \( {\nel}_i{\nel}_i^{\top} \) and its noise-reduced counterpart \( \nel_i'\nel_i'^{\top} \) are retained in \eqrefp{eq:K1}.



\textbf{Difference From Averaging Prior to Optimization:} Next, we show that such gradient averaging differs from the averaging operation conducted prior to optimization. Let \( \vec{G}' = [{\nel}'_1, {\nel}'_2, \ldots, {\nel}'_n]^{\top} \) such that $\vec{G}'$ represents the row-wise convex combination of \( \vec{G} \). Approximating the Hessian of the \gls{miqp} \eqrefp{eq:miqp}, two scenarios emerge: using \( \vec{G} \) that not performing averaging (case 1) and \( \vec{G}' \) that performs averaging before optimization (case 2).

\textcolor{darkgreen}{\textbf{Case 1}} is the original \gls{lr} formulation~\eqrefp{eq:Euclidean_formulation}. Denote by $K$ below its term corresponding to $K^{(1)}_{\vec{\Pi}}$:
\begin{align}
    K & =(\vec{w}-\vec{\bw})^\top\vec{G}^{\top}\vec{G}(\vec{w}-\vec{\bw}) \nonumber \\
     & = \sum_{i=1}^n(\vec{w}^\top{\nel}_i{{\nel}}_i^\top\vec{w}-\vec{w}^\top{\nel}_i{{\nel}}_i^\top\vec{\bw}-\vec{\bw}^\top{\nel}_i{{\nel}}_i^\top\vec{w}+\vec{\bw}^\top{\nel}_i{{\nel}}_i^\top\vec{\bw})
\label{eq:K}
\end{align}
\textcolor{dtured}{\textbf{Case 2}} uses the less-noisy row-wise convex combination matrix $\vec{G}'$ instead of $\vec{G}$. Yet, the original covariance ${\nel}_i{\nel}_i^{\top}$ is lost: Denote by $K'$ the corresponding term, and we have
\begin{equation}
K'  = \sum_{i=1}^n(\vec{w}^\top\nel_i'{{{\nel}'}}_i^\top\vec{w}-\vec{w}^\top\nel_i'{{{\nel}'}}_i^\top\vec{\bw}-\vec{\bw}^\top\nel_i'{{{\nel}'}}_i^\top\vec{w}+\vec{\bw}^\top\nel_i'{{{\nel}'}}_i^\top\vec{\bw})
\label{eq:K'} %
\end{equation}

Inspecting the expressions, it can be observed that $K_{\vec{\Pi}}^{(1)}$ (also $K_{\vec{\Pi}}^{(2)}$) strikes a balance between $K$ and $K'$. There are two notable extreme cases for \(\vec{\Pi}\): 
\begin{enumerate}
    \item \(\vec{\Pi}=\diag(1/n)\). This corresponds to the \gls{lr} formulation, as detailed in Section~\ref{sec:formulation}. A smaller value of \(\varepsilon\) steers the optimization in this direction.
    \item \(\vec{\Pi}=(1/n^2)\vec{1}\cdot\vec{1}^{\top}\). This arises when \(\varepsilon\rightarrow\infty\), meaning the entropy term holds sway in the \gls{ot} optimization. Here, mutual information is minimized to ensure an even contribution from data points in the convex combination. Both \(x'_i\) and \(y'_i\) are the arithmetic means of their respective sets, and all \(\nel_i'\) are equivalent to the averaged gradient over the \(n\) points. Importantly, the original covariance remains intact even in this edge case.
\end{enumerate}

As \(n\) grows indefinitely, the empirical \gls{ot} formulation from \eqrefp{eq:wasserstein_formulation_empirical} approaches its continuous counterpart given by \eqrefp{eq:wasserstein_formulation}. Intuitively, a large dataset of high-quality training samples makes the empirical fisher a close approximation to the true fisher. In such situations, \(\varepsilon\) is set to zero. Brenier's theorem \citep{peyre2019course} then suggests that the \gls{ot} plan turns into a monotone map for costs represented by squared Euclidean distances. This means \(\vec{\Pi}\) tends towards \(\diag(1/n)\). Consequently, the Wasserstein distance formulation reduces to the Euclidean distance formulation, delivering optimal performance with ample data.

An advantage of employing the \gls{ewr} formulation is its inherent capability of gradient averaging. This approach negates the need to manually determine the convex combination coefficients or resort to density estimation to pinpoint the nearest gradient neighbors for averaging.  Importantly, this seamless trade-off has an advantage over using Euclidean distance with gradient averaging performed prior to optimization. The reason is that the original covariance information will inevitably be lost in the formulation \eqrefp{eq:K'}, irrespective of the chosen averaging method.

\textbf{Sample Compexity} of $W_2^2(\mu, \nu)$ is narrowed to $O(1/\sqrt{n})$ from $O(1/n^{\frac{1}{4}})$ by the entropic regularization term. 
Please see Appendix~\ref{sec:sample_complexity} for details.



\section{Algorithm Design}

\textbf{Algorithmic Framework}. The algorithm addresses the network pruning problem defined in \eqrefp{eq:wasserstein_formulation}. Drawing inspiration from \citep{chen2022network}, the algorithm incrementally adjusts the sparsity of the weights vector $\vec{w}$ by using a descending sequence of non-zero elements $k_0,\ldots,k_T$. During each sparsity level, the weights $\vec{w}$ and the transportation plan $\vec{\Pi}$ (can be obtained with efficient algorithms; see Appendix \ref{sec:ot-algorithms}) are refined iteratively.

\begin{algorithm}
\caption{\texttt{S}parse Entropic \texttt{WA}sserstein Regression \texttt{P}runing (\texttt{SWAP})}
\label{alg:cap}
\begin{algorithmic}[1]
\Require{Number of pruning stages $T$, pre-pruning weights $\vec{\bw}$, target sparsity $k$, regularization parameter $\lambda$, $\varepsilon$, batches $\B_0,\B_1\ldots,\B_T$, optimization step size $\tau>0$.}
\Ensure{Post-pruning weights $\vec{w}$, satisfying $\norm{\vec{w}}\leq k$}
\State Set $k_0,k_1,\ldots,k_T$ as a descending sequence, with $k_0<p$ and $k_T=k$.
\State $\vec{w}^{(0)}\gets\vec{\bw}$
\For{$t \gets 0, 1,\ldots,T$}
\State Compute $\vec{G}=[\nel_1(\vec{\bw}),\ldots \nel_n(\vec{\bw})]^{\top}$ with batch $\B_t$ \label{alg:cap-G}
\State $\vec{x},\vec{y}\gets\vec{G}\vec{w}^{(t)}, \vec{G}\vec{\bw}$ 
\State Compute the pairwise Euclidean distance matrix $\vec{C}$ between $\vec{x}$ and $\vec{y}$
\State Compute \gls{ot} planning $\vec{\Pi}^{(t)}$ (see \appendixname~\ref{sec:ot-algorithms}) \label{alg:cap-ot}
\State $\nabla Q \gets \vec{G}^\top (\vec{\Pi} (\vec{G}\vec{w}^{(t)} - \vec{G}\vec{\bw})) + \lambda(\vec{w}^{(t)} - \vec{\bw})$
\State $\vec{w}^{(t+\frac{1}{2})}\gets \vec{w}^{(t)}-\tau\nabla Q$ \label{alg:cap-sgd}
\State $\vec{w}^{(t+1)}\gets \text{Select from $\vec{w}^{(t+\frac{1}{2})}$ $k_t$ components having largest absolute values; \label{alg:cap-iht}
Others zero }$
\State $\vec{\bw}\gets \vec{w}^{(t+1)}$ \label{alg:cap-bw}
\EndFor
\State $\vec{w} = \vec{w}^{(T+1)}$
\end{algorithmic}
\end{algorithm}

\textbf{Weights Optimization}. 
The weights $\vec{w}$ are optimized using the \gls{sgd} paired with the \gls{iht} algorithm. We use $\nabla Q$ to represent the derivative of $Q(\vec{w})$ for brevity, with its comprehensive derivation in Appendix~\ref{sec:Q(w)}. The expression is
\begin{equation}
\nabla Q = \vec{G}^\top (\vec{\Pi} (\vec{G}\vec{w} - \vec{G}\vec{\bw})) + \lambda(\vec{w} - \vec{\bw}).
\end{equation}

Following the weight updates driven by \gls{sgd} (as seen in line~\ref{alg:cap-sgd} of Algorithm~\ref{alg:cap}), the \gls{iht} method is applied. Here, the $k_t$ components of $\vec{w}$ with the largest magnitudes are retained, while the remaining are set to zero, ensuring adherence to the sparsity criteria.

A vital component of the optimization process is the choice of the stepsize $\tau$ (referenced in line~\ref{alg:cap-sgd}). Although a straightforward approach might be to set $\tau = \frac{1}{L}$ (where $L$ denotes the Lipschitz constant of $Q$), better performance can be achieved when the stepsize is optimized using the methodology proposed in \citep[Algorithm~2]{chen2022network}. For the quadratic function $Q$, the Lipschitz constant $L$ is given by $L = n\lambda + \norm{G}^2_{op}$, where $\norm{\cdot}_{op}$ indicates the foremost singular value.

Line~\ref{alg:cap-iht} in Algorithm~\ref{alg:cap} employs the \gls{iht} method that is commonly used in sparse learning, which together with line \ref{alg:cap-sgd}, forms a projected gradient descent algorithm. It finds a sparse representation of the updated gradient in line \ref{alg:cap-sgd}. Intuitively, \gls{iht} keeps the dominant model weights and essentially preserves the most impactful aspects of the to-be-trimmed model. Although there exist alternative strategies for refining the \gls{iht} solution---including active set updates, coordinate descent, and the Woodbury formula's back-solving---a discussion on these falls outside the scope of this paper. For in-depth exploration, especially with respect to the specialized case described in \eqrefp{eq:special_case}, one can consult \citep{bhatia2015robust,chen2020multivariate,hazimeh2020fast,benbaki2023fast}.

\section{Numerical Results}

\begin{table}[!h]
    \centering
    \caption{Model Pruning Accuracy Benchmarking. Five runs are taken for LR and EWR, with the mean and 95\% confidence interval (in the brackets) reported. The data of MP, WF, and CBS are copied from \citep{yu2022combinatorial}. The superscript ``$+\sigma$'' indicates that 20\% of data is with noise. Bold texts imply the best performance, with $0.1$ percentage as tolerance in difference. The sparsity column is the target sparsity. }
    \label{tab:accuracy}
    \begin{tabular}{c|c|ccccc}
    \toprule
       Network & Sparsity & MP & WF & CBS & LR (EWR$_{\vec{\Pi}=\vec{I}/n}$) & EWR (proposed) \\ 
    \midrule
        \multirow{9}{*}{\shortstack{MLPNet \\ on MNIST \\ (93.97\%)}} & 0.5 & 93.93 & 94.02 & 93.96 & \textbf{95.26} ($\pm$0.03) & \textbf{95.24} ($\pm$0.03) \\ 
         & 0.6 & 93.78 & 93.82 & 93.96 & \textbf{95.13} ($\pm$0.02) & \textbf{95.13} ($\pm$0.01) \\ 
         & 0.7 & 93.62 & 93.77 & 93.98 & \textbf{94.93} ($\pm$0.03) & \textbf{95.05} ($\pm$0.04) \\ 
         & 0.8 & 92.89 & 93.57 & 93.90 & \textbf{94.82} ($\pm$0.04) & \textbf{94.84} ($\pm$0.03) \\ 
         & 0.9 & 90.30 & 91.69 & 93.14 & \textbf{94.32} ($\pm$0.05) & \textbf{94.30} ($\pm$0.05) \\ 
         & 0.95 & 83.64 & 85.54 & 88.92 & \textbf{92.82} ($\pm$0.06) & \textbf{92.86} ($\pm$0.05) \\ 
         & 0.95$^{+\sigma}$ & - & - & - & 90.11 ($\pm$0.08) & \textbf{90.50} ($\pm$0.07) \\ 
         & 0.98 & 32.25 & 38.26 & 55.45 & 84.43 ($\pm$0.10) & \textbf{85.71} ($\pm$0.09) \\ 
         & 0.98$^{+\sigma}$ & - & - & - & 82.12 ($\pm$0.11) & \textbf{83.69} ($\pm$0.10) \\ 
    \midrule
        \multirow{9}{*}{\shortstack{ResNet20 \\ on CIFAR10 \\ (91.36\%)}} & 0.5 & 88.44 & 90.23 & 90.58 & \textbf{92.06} ($\pm$0.04) & \textbf{92.04} ($\pm$0.03) \\ 
         & 0.6 & 85.24 & 87.96 & 88.88 & \textbf{91.98} ($\pm$0.09) & \textbf{91.98} ($\pm$0.09) \\ 
         & 0.7 & 78.79 & 81.05 & 81.84 & 91.09 ($\pm$0.10) & \textbf{91.89} ($\pm$0.10) \\ 
         & 0.8 & 54.01 & 62.63 & 51.28 & 89.00 ($\pm$0.12) & \textbf{90.15} ($\pm$0.09) \\ 
         & 0.9 & 11.79 & 11.49 & 13.68 & 87.63 ($\pm$0.11) & \textbf{88.82} ($\pm$0.10) \\ 
         & 0.95 & - & - & - & 80.25 ($\pm$0.17) & \textbf{81.33} ($\pm$0.15) \\
         & 0.95$^{+\sigma}$ & - & - & - & 77.37 ($\pm$0.18) & \textbf{79.05} ($\pm$0.16) \\
         & 0.98 & - & - & - & 68.15 ($\pm$0.27) & \textbf{69.21} ($\pm$0.24) \\
         & 0.98$^{+\sigma}$ & - & - & - & 65.04 ($\pm$0.27) & \textbf{68.01} ($\pm$0.25) \\
    \midrule 
        \multirow{4}{*}{\shortstack{ResNet50 \\ on CIFAR10 \\ (92.78\%)}}
         & 0.95 & - & - & - & 83.75 ($\pm$0.14) & \textbf{84.96} ($\pm$0.15) \\
         & 0.95$^{+\sigma}$ & - & - & - & 82.34 ($\pm$0.16) & \textbf{84.92} ($\pm$0.17) \\
         & 0.98 & - & - & - & 81.04 ($\pm$0.14) & \textbf{82.85} ($\pm$0.20) \\
         & 0.98$^{+\sigma}$ & - & - & - & 80.11 ($\pm$0.23) & \textbf{82.94} ($\pm$0.22) \\
    \midrule
        \multirow{9}{*}{\shortstack{MobileNetV1 \\ on ImageNet \\ (71.95\%)}} & 0.3 & 71.60 & \textbf{71.88} & \textbf{71.87}  & 71.14 ($\pm$0.08) & \textbf{71.87} ($\pm$0.05)\\ 
         & 0.4 & 69.16 & 71.15 & \textbf{71.45} & 71.12 ($\pm$0.10) &  \textbf{71.44} ($\pm$0.07)\\ 
         & 0.5 & 62.61 & 68.91 & 70.21 & 70.12 ($\pm$0.13) &  \textbf{71.12} ($\pm$0.18)\\ 
         & 0.6 & 41.94 & 60.90 & 66.37 & 70.05 ($\pm$0.22) &  \textbf{70.92} ($\pm$0.18)\\ 
         & 0.7 & 6.78 & 29.36 & 55.11 & 68.15 ($\pm$0.17) & \textbf{69.26} ($\pm$0.13) \\ 
         & 0.8 & 0.11 & 0.24 & 16.38 & 65.72 ($\pm$0.19) & \textbf{66.82} ($\pm$0.14) \\ 
         & 0.8$^{+\sigma}$ & - & - & - & 60.29 ($\pm$0.18) & \textbf{63.62} ($\pm$0.15) \\ 
         & 0.9 & - & - & - & 47.65 ($\pm$0.15) & \textbf{49.43} ($\pm$0.13) \\
         & 0.9$^{+\sigma}$ & - & - & - & 44.55 ($\pm$0.16) & \textbf{47.98} ($\pm$0.16) \\
    \bottomrule
    \end{tabular}
\end{table}

Our method is compared with several existing SoTA methods including MP (magnitude pruning \citep{mozer1989using}), WF (WoodFisher \citep{singh2020woodfisher}), CBS (Combinatorial Brain Surgeon \citep{yu2022combinatorial}), and LR (i.e. the sparse \gls{lr} formulation adopted by \citep{chen2022network}). We refer to our proposed method as EWR (i.e. sparse entropic Wasserstein regression). Note that LR is a special instance of EWR, with $\vec{\Pi}=\diag(1/n)$. All the methods are benchmarked on pre-trained neural networks: MLPNet (30K parameters) trained on MNIST \citep{lecun1998gradient}, ResNet20  (200K parameters) and ResNet50 (25M parameters) \citep{he2016deep} trained on CIFAR10 \citep{krizhevsky2009learning}, and MobileNetV1 \citep{howard2017mobilenets} (4.2M parameters) trained on ImageNet \citep{deng2009imagenet}. The experiment setup for reproducibility\footnote{The code is available on \faGithub \\ \url{https://github.com/youlei202/Entropic-Wasserstein-Pruning}} is detailed in \appendixname\ref{sec:experiment_setup}. We deliver more experiments results in \appendixname~\ref{sec:ablation_study}-\ref{sec:ot_vis}.

\textbf{Model Accuracy Performance Benchmarking}. \tablename~\ref{tab:accuracy} compares different networks across various sparsity levels, utilizing different methods. MLPNet's performance on MNIST is consistent across different sparsity levels, with both LR and the proposed EWR method showing superior performance. The advantages of EWR over the others are reflected by the three more challenging tasks ResNet20 and ResNet50 on CIFAR10 and MobileNetV1 on ImageNet, especially in the presence of noisy gradients. In summary, the proposed EWR method consistently outperforms or matches other methods. The LR method performs well at lower sparsity levels but is surpassed otherwise.
\begin{table}[!h]
\centering
\caption{Comparison of testing loss values (no fine-tuning) for ResNet20. The result is averaged over 25 runs. The 90\% confidence interval is reported. The target sparsity is set to be 0.95.}
\label{tab:loss_resnet20}
\begin{tabular}{c|*{4}{ccc}}
\toprule
& \multicolumn{6}{c}{10\% Noisy Data} \\
  Sparsity       & \multicolumn{3}{c}{Noise = \(\sigma\)} & \multicolumn{3}{c}{Noise = \(2\sigma\)} \\
\cmidrule(lr){2-4} \cmidrule(lr){5-7}
      & LR   & EWR  & Diff & LR   & EWR  & Diff \\
\midrule
0.95 & 2.83 ($\pm$0.02) & \textbf{2.75} ($\pm$0.02)  & \textcolor{teal}{\textbf{2.87}\%} & 2.86 ($\pm$0.02)  & \textbf{2.74} ($\pm$0.01) & \textcolor{teal}{\textbf{4.35}\%} \\
0.84 & 1.58 ($\pm$0.01)  & \textbf{1.54} ($\pm$0.01) & \textcolor{teal}{\textbf{2.73}\%} & 1.62 ($\pm$0.03)  & \textbf{1.54} ($\pm$0.02)  & \textcolor{teal}{\textbf{5.15}\%} \\
0.74 & 0.66 ($\pm$0.00) & \textbf{0.64} ($\pm$0.00) & \textcolor{teal}{\textbf{3.03}\%} & 0.66 ($\pm$0.00)  & \textbf{0.65} ($\pm$0.00)  & \textcolor{teal}{\textbf{1.32}\%} \\
0.63 & 0.35 ($\pm$0.00)  & 0.35 ($\pm$0.00)         & 0.00\% & 0.35 ($\pm$0.00) & 0.35 ($\pm$0.00)         & 0.85\% \\
\bottomrule
\end{tabular}
\begin{tabular}{c|*{4}{ccc}}
& \multicolumn{6}{c}{25\% Noisy Data} \\
  Sparsity       & \multicolumn{3}{c}{Noise = \(\sigma\)} & \multicolumn{3}{c}{Noise = \(2\sigma\)} \\
\cmidrule(lr){2-4} \cmidrule(lr){5-7}
      & LR   & EWR  & Diff & LR   & EWR & Diff \\
\midrule
0.95 & 2.87 ($\pm$0.02) & \textbf{2.77} ($\pm$0.01) & \textcolor{teal}{\textbf{3.50}\%} & 2.89 ($\pm$0.02) & \textbf{2.79} ($\pm$0.02) & \textcolor{teal}{\textbf{3.82}\%} \\
0.84 & 1.72 ($\pm$0.02) & \textbf{1.65} ($\pm$0.02) & \textcolor{teal}{\textbf{4.07}\%} & 1.76 ($\pm$0.02) & \textbf{1.69} ($\pm$0.02) & \textcolor{teal}{\textbf{4.05}\%} \\
0.74 & 0.67 ($\pm$0.01) & 0.67 ($\pm$0.00)         & 0.49\% & 0.68 ($\pm$0.00) & \textbf{0.67} ($\pm$0.00) & \textcolor{teal}{\textbf{1.55}\%} \\
0.63 & 0.36 ($\pm$0.00) & 0.36 ($\pm$0.00)         & 0.00\% & 0.35 ($\pm$0.00) & 0.35 ($\pm$0.00)          & 0.00\% \\
\bottomrule
\end{tabular}
\end{table}
\begin{table}[!h]
\centering
\caption{Comparison of testing loss values (no fine-tuning) for MobileNetV1. The result is averaged from 10 runs. The 90\% confidence interval is reported. The target sparsity is set to be 0.75.}
\label{tab:loss_mobilenetv1}
\begin{tabular}{c|*{4}{ccc}}
\toprule
& \multicolumn{6}{c}{10\% Noisy Data} \\
  Sparsity       & \multicolumn{3}{c}{Noise = \(\sigma\)} & \multicolumn{3}{c}{Noise = \(2\sigma\)} \\
\cmidrule(lr){2-4} \cmidrule(lr){5-7}
      & LR   & EWR  & Diff & LR   & EWR  & Diff \\
\midrule
0.75 &   4.54 ($\pm$0.06)   &  \textbf{4.34} ($\pm$0.06)   & \textcolor{teal}{\textbf{4.41}\%}  &   4.62 ($\pm$0.07)  &   \textbf{4.40} ($\pm$0.06)  &  \textcolor{teal}{\textbf{5.02}\%}   \\
0.63 &   2.53 ($\pm$0.04)  &  \textbf{2.46} ($\pm$0.03)   & \textcolor{teal}{\textbf{2.61}\%}  &   2.56 ($\pm$0.04)  &   \textbf{2.48} ($\pm$0.04)  &  \textcolor{teal}{\textbf{3.23}\%}       \\
0.53 &   1.64 ($\pm$0.02)  &  \textbf{1.56} ($\pm$0.02)   & \textcolor{teal}{\textbf{5.16}\%}  &   1.66 ($\pm$0.03)  &   \textbf{1.56} ($\pm$0.02) &    \textcolor{teal}{\textbf{6.01}\%}       \\
0.42 &   1.30 ($\pm$0.00)  &  1.30 ($\pm$0.00)   &    0.00\%       &   1.30 ($\pm$0.00)  &   1.30 ($\pm$0.00)  &      0.00\%     \\
\bottomrule
\end{tabular}
\begin{tabular}{c|*{4}{ccc}}
& \multicolumn{6}{c}{25\% Noisy Data} \\
  Sparsity       & \multicolumn{3}{c}{Noise = \(\sigma\)} & \multicolumn{3}{c}{Noise = \(2\sigma\)} \\
\cmidrule(lr){2-4} \cmidrule(lr){5-7}
      & LR   & EWR  & Diff & LR   & EWR  & Diff \\
\midrule
0.75 &   4.93 ($\pm$0.06)  &  \textbf{4.53} ($\pm$0.06)   & \textcolor{teal}{\textbf{8.13}\%}  &   5.02  ($\pm$0.06) &   \textbf{4.66} ($\pm$0.05)  &     \textcolor{teal}{\textbf{7.92}\%}       \\
0.63 &   2.54 ($\pm$0.04)   &  \textbf{2.44} ($\pm$0.03)    & \textcolor{teal}{\textbf{4.01}\%}  &   2.57 ($\pm$0.04)   &   \textbf{2.45} ($\pm$0.03)   &    \textcolor{teal}{\textbf{5.00}\%}       \\
0.53 &   1.66 ($\pm$0.02)   &  \textbf{1.57} ($\pm$0.02)    & \textcolor{teal}{\textbf{5.16}\%}  &   1.66 ($\pm$0.02)   &   \textbf{1.55} ($\pm$0.02)   &       \textcolor{teal}{\textbf{6.63}\%}     \\
0.42 &   1.30 ($\pm$0.00)   &  1.30 ($\pm$0.00)    &   0.30\%        &   1.31 ($\pm$0.00)    &  1.31 ($\pm$0.00)     &    0.00\%       \\
\bottomrule
\end{tabular}
\end{table}

\textbf{Robustness with Noisy Gradient}. From Section~\ref{sec:theoretical}, EWR differs from LR in terms of gradient noise reduction achieved by solving the \gls{ot} problem to obtain a group of non-trivial data pair weighting coefficients. Hence, LR that has the transportation plan $\vec{\Pi}$ fixed to $\diag(1/n)$ naturally serves as a baseline for evaluating the effectiveness of such optimization in terms of robustness against noise.
In two noisy scenarios, 10\%, and 25\%, we evaluate loss at noise levels of $\sigma$ and $2\sigma$ across varying sparsity. \tablename{s}~\ref{tab:loss_resnet20} and \ref{tab:loss_mobilenetv1} contrast the loss difference between LR and EWR. EWR consistently outperforms LR in both ResNet20 and MobileNetV1, most notably in noisy conditions and at higher sparsity. The peak performance difference is 8.13\% favoring EWR on MobileNetV1 at 0.75 sparsity with 25\% noise. Hence, EWR outperforms LR.

\section{Conclusions and Future Impact}

The paper offers a novel formulation based on \gls{ewr}, which strikes a balance between covariance information preservation and noise reduction. 
The work suggested promising avenues for applications in large-scale model compression, though it may require further empirical validation and exploration of practical implementations.

\bibliography{iclr2024_conference}
\bibliographystyle{iclr2024_conference}

\appendix

\section{Appendix}

\subsection{Proof of Proposition~\ref{thm:convex_equality}}
\label{sec:proof_convex_equality}
\textit{(Convex Hull Distance Equality)}
Consider a set $S$ and its convex hull $\textit{Conv}(S)$ in a Euclidean space, and an arbitrary point ${x}$ in the space. For any probability measure $\hat{\nu}$ on $S$, we can find a point ${y}'$ in $\textit{Conv}(S)$ as ${y}' = \int {y} \, \mathrm{d}\nu({y})$ such that $\|{x} - {y}'\|^2 = \int \|{x} - {y}\|^2 \, \mathrm{d}\hat{\nu}({y})$, where $\nu$ is a measure on $\textit{Conv}(S)$.
\begin{proof}
We define a function $f(\nu) = \norm{{x} - {y}'}^2$, where ${y}' = \int {y} \mathrm{d}\nu({y})$. This function takes the empirical measure $\nu$ as input and computes the squared Euclidean distance between ${x}$ and ${y}'$. Similarly, we define a function $g(\nu) = \int \norm{{x} - {y}}^2\mathrm{d}\nu({y})$, which computes the weighted average of squared Euclidean distances between ${x}$ and the points in the set $S$ according to the probability measure $\nu$.

Without loss of generality, let's assume that $S$ is contained within its convex hull $\textit{Conv}(S)$. Then, the right-hand side of the equation to be proved in the theorem takes the minimum and maximum values, respectively, at points within $\textit{Conv}(S)$, i.e.,
\[
g_{\min} = \inf_{\nu} \int \|{x}-{y}\|^2 \, \mathrm{d}\nu({y})
\]
and
\[
g_{\max} = \sup_{\nu} \int \|{x}-{y}\|^2 \, \mathrm{d}\nu({y}).
\]

The function $f(\nu)$ takes its maximum value at ${z}$, where ${z}$ is the farthest point inside $\textit{Conv}(S)$ from ${x}$, i.e.,
\[
f_{\max} = \sup_{\nu} \|{x}-{y}'\|^2 = \|{x}-{z}\|^2.
\]
Similarly, the minimum value is obtained at ${y}'={z}'$, where ${z}'$ is the closest point inside $\textit{Conv}(S)$ to ${x}$. The minimum value can reach zero if ${x}$ is inside $\textit{Conv}(S)$. Formally,
\[
f_{\min} = \inf_{\mu} \left\{ \|{x}-{z}\|^2 \middle|~{z}\in \textit{Conv}(S) \right\}.
\]

To establish that $f_{\max} \geq g_{\max}$, we consider the maximum values of the two functions. The function $f(\nu)$ takes its maximum value at ${z}$, which is the farthest point inside $\textit{Conv}(S)$ from ${x}$. This means that $f_{\max}$ is the squared Euclidean distance between ${x}$ and ${z}$, i.e., $f_{\max} = |{x}-{z}|^2$. On the other hand, the function $g(\nu)$ computes the weighted average of squared Euclidean distances between ${x}$ and the points in $S$. The maximum value of $g(\nu)$, denoted as $g_{\max}$, corresponds to the squared Euclidean distance between ${x}$ and the farthest point in $S$. Since $\textit{Conv}(S)$ contains $S$, it follows that ${z}$ is farther from ${x}$ than any point in $S$. Therefore, $f_{\max} \geq g_{\max}$. Similarly, since $\textit{Conv}(S)$ contains $S$ and ${z}'$ is the closest point in $\textit{Conv}(S)$ to ${x}$, it follows that $f_{\min} \leq g_{\min}$.

We apply the intermediate value theorem to finish the proof. For any measure $\nu$, let $h(\mu)=f(\mu)-g(\nu)$. Once can find two measures $\mu_1$ and $\mu_2$ such that $h(\mu_1)\leq 0$ and $h(\mu_2)\geq 0$, hence the zero point of $h$ exists, followed by that there is a corresponding value of $\mu$ such that $f(\mu) = g(\nu)$.

Hence, the conclusion.
\end{proof}


\subsection{Sample Complexity}
\label{sec:sample_complexity}
The efficiency of an estimator is often measured by its ability to deliver accurate estimates with fewer samples, a trait referred to as 'good sample complexity'.  The entropic Wasserstein distance formulation is posited to enhance noise reduction by optimizing this sample complexity. When \(\varepsilon = 0\), the sample complexity of \(\ot(\mu,\nu)\) stands at \(O(1/n^{\frac{1}{4}})\). Specifically, 
\[
\lim_{n\rightarrow \infty}\EE[\ot(\mu_n, \nu_n) - \ot(\mu, \nu)] = O(1/n^{\frac{1}{4}})
\]
as per \citep[Corollary 2]{nadjahi2020statistical}. As \(\varepsilon\) approaches infinity, \(\ot(\mu, \nu)\) leverages the beneficial characteristics of the \gls{mmd} \citep{gretton2006kernel}, which narrows the sample complexity to \(O(1/\sqrt{n})\) according to \citep[Theorem 3]{genevay2019sample}. The coefficient \(\varepsilon\) acts as a regulator, adjusting the sample complexity within this specified range.

\subsection{\gls{ot} Plan Optimization}
\label{sec:ot-algorithms}

For an arbitrary $\vec{w}^{(t)}$, the transportation plan $\vec{\Pi}^{(t)}$ is obtained by solving $\ot(x(\vec{w}^{(t)}),y)$. The optimization of the transportation plan $\vec{\Pi}$ in line~\ref{alg:cap-ot} of Algorithm~\ref{alg:cap} is based on the Sinkhorn-Knopp algorithm, shown in Algorithm~\ref{alg:sinkhorn} below.
\begin{algorithm}
\caption{Sinkhorn-Knopp Algorithm for Regularized OT}\label{alg:sinkhorn}
\begin{algorithmic}[1]
\Require{$\vec{C}$ ($n\times n$ Euclidean distance matrix), $\bm{\mu}, \bm{\nu}$ (probability mass of $\vec{x}$ and $\vec{y}$), $\varepsilon$ (regularization parameter), $\epsilon$ (tolerance for stopping criterion)}
\Ensure{$\vec{\Pi}$}
\State Initialize $\vec{K} = \exp(-\vec{C}/\varepsilon)$, $\vec{u} = \vec{1}_n/n$, $\vec{v} = \vec{1}_n/n$
\Repeat
  \State $\vec{u} = \nicefrac{\bm{\mu}}{\vec{Kv}}$
  \State $\vec{v} = \nicefrac{\bm{\nu}}{\vec{K}^\top\vec{u}}$
\Until{$\sup\left\{\norm{\vec{a} - \diag (\vec{u})\vec{Kv}}_\infty, \norm{\vec{b} - \diag (\vec{v})\vec{K}^\top\vec{u}}_\infty\right\} < \epsilon$}
\State \Return $\vec{\Pi} = \diag(\vec{u})\cdot\vec{K}\cdot\diag(\vec{v})$
\end{algorithmic}
\end{algorithm}

The Sinkhorn-Knopp algorithm is an iterative method used to solve regularized \gls{ot} problems, aiming to find a transportation plan that minimizes total cost while adhering to specific source and target probability distributions. In the algorithm, an initial matrix $\vec{K}$ is formed using the exponential of the negative cost matrix divided by a regularization parameter, and two vectors are initialized as uniform distributions. These vectors are then iteratively updated using rules derived from the Kullback-Leibler divergence, seeking to align the row and column sums of the resulting matrix with the given source and target distributions. The process continues until the maximum difference between the actual and desired row and column sums falls below a specified tolerance, ensuring the solution is feasible.

Note that $x$ and $y$ are one-dimensional projections of high-dimensional random variables $\nabla \ell$, of which the dimension is the number of parameters of the neural network, a.k.a. $p$. Prior research has demonstrated that when certain moderate criteria are met, the distribution of lower-dimensional versions of high-dimensional random variables tends to closely follow a Gaussian (or normal) distribution \citep{sudakov1978typical, reeves2017conditional, nadjahi2021fast}.
In the research by \citep{janati2020entropic}, it is highlighted that given $x$ and $y$ obeying the Gaussian distribution, $\ot(x,y)$ can be reduced to a concise closed-form solution. As illustrated in Algorithm~\ref{alg:ot-closed-form}, the \gls{ot} plan can be directly computed relying solely on the statistics of the empirical distributions $\vec{x}$ and $\vec{y}$. 
\begin{algorithm}
\caption{Closed Form Algorithm for Regularized OT}\label{alg:ot-closed-form}
\begin{algorithmic}[1]
\Require{$\psi = \sqrt{\frac{\varepsilon}{2}}$, $a=\text{mean}(\vec{x})$, $b=\text{mean}(\vec{y})$, $\sigma^2_a=\text{var}(\vec{x})$, $\sigma^2_b=\text{var}(\vec{y})$, $d_\psi = (4\sigma_a \sigma_b^2 \sigma_a + \psi^4)^{\frac{1}{2}}$}
\Ensure{
\[
\vec{\Pi} \sim \N\left(
\begin{bmatrix}
a \\
b 
\end{bmatrix}, 
\begin{bmatrix}
    \sigma_a^2 & \frac{1}{2}\sigma_a d_\psi\sigma_a^{-1} \\
    \frac{1}{2}\sigma_a d_\psi\sigma_a^{-1} & \sigma_b^2
\end{bmatrix}
\right)
\]
}
\end{algorithmic}
\end{algorithm}

Specifically, the \gls{ot} plan, represented as $\vec{\Pi}$, is governed by the mean and variance of both $x$ and $y$, in tandem with the regularization parameter $\varepsilon$. To derive the $n\times n$ matrix $\vec{\Pi}$, $(x_i,y_i)$, where $i$ spans from 1 to $n$, are employed to extract $n$ samples from the distribution produced in Algorithm~\ref{alg:ot-closed-form}.

\begin{figure}[t]
\centering
\begin{tikzpicture}
\begin{axis}[
  title={Sinkhorn-Knopp vs Closed-Form Solution},
  legend cell align={left},
  xlabel={$\varepsilon$},
   ylabel style={align=center},
  ylabel={ \gls{ot} Objective $\ot$\\ (Without Regularization Term)},
  legend pos = {north west},
  scaled ticks=false,
  tick label style={/pgf/number format/fixed},
grid=both,
minor tick num=1,
grid style={line width=.1pt, dashed, draw=black!10},
  ]
  
\addplot[blue, mark=none] coordinates {
(1,1005.802952042648)
(2,1006.0639072438831)
(3,1006.305483685531)
(4,1006.5499414683597)
(5,1006.8043732864627)
(6,1007.0659125098668)
(7,1007.3335352674474)
(8,1007.6058261776282)
(9,1007.8826876702506)
(10,1008.1638648075632)
(11,1008.4474094394752)
(12,1008.7314523149624)
(13,1009.0151327143907)
(14,1009.2988201477501)
(15,1009.5827172124161)
(16,1009.8658857160771)
(17,1010.1476107637914)
(18,1010.427632256389)
(19,1010.7056675734411)
(20,1010.981524866781)
(21,1011.2552293263243)
(22,1011.5269435702455)
(23,1011.7968882890796)
(24,1012.0653072265575)
(25,1012.3324490060611)
(26,1012.5985548118446)
(27,1012.8638499399697)
(28,1013.1285384892412)
(29,1013.3928005521605)
(30,1013.6567913105871)
(31,1013.9206415049459)
(32,1014.1844588216809)
(33,1014.4483298249884)
(34,1014.7123221407167)
(35,1014.9764866782417)
(36,1015.2408597459207)
(37,1015.50546497402)
(38,1015.7703150039679)
(39,1016.0354129346945)
(40,1016.3007535374704)
(41,1016.5663242635112)
(42,1016.8321060773917)
(43,1017.0980741583295)
(44,1017.3641985240376)
(45,1017.630444650769)
(46,1017.8967741874872)
(47,1018.1631458855068)
(48,1018.4295168716055)
(49,1018.695844356161)
(50,1018.9620877579508)
(51,1019.2282110300766)
(52,1019.4941847238687)
(53,1019.7599871408231)
(54,1020.0256039549068)
(55,1020.291026042182)
(56,1020.5562458565452)
(57,1020.8212532687173)
(58,1021.0860320279114)
(59,1021.3505577780707)
(60,1021.6147979953329)
(61,1021.8787135962987)
(62,1022.1422615483673)
(63,1022.4053976970359)
(64,1022.6680791697964)
(65,1022.9302660019073)
(66,1023.1919219252159)
(67,1023.4530144746486)
(68,1023.7135146625786)
(69,1023.9733964645359)
(70,1024.2326362935355)
(71,1024.491212559213)
(72,1024.7491053418019)
(73,1025.0062961708823)
(74,1025.2627678824906)
(75,1025.5185045271314)
(76,1025.773491307193)
(77,1026.0277145295736)
(78,1026.28116156531)
(79,1026.5338208119742)
(80,1026.7856816569147)
(81,1027.0367344406177)
(82,1027.2869704200107)
(83,1027.536381731748)
(84,1027.7849613556057)
(85,1028.0327030781473)
(86,1028.2796014567798)
(87,1028.5256517843382)
(88,1028.770850054325)
(89,1029.015192926879)
(90,1029.2586776955832)
(91,1029.5013022551686)
(92,1029.7430650701845)
(93,1029.983965144674)
(94,1030.2240019929015)
(95,1030.4631756111573)
(96,1030.701486450649)
(97,1030.9389353915014)
(98,1031.175523717851)
(99,1031.4112530940495)
(100,1031.6461255419356)
};
\addlegendentry{Sinkhorn-Knopp}

\addplot[red, mark=none] coordinates {
(1,1005.408935546875)
(2,1008.3507080078125)
(3,1011.031494140625)
(4,1013.5426635742188)
(5,1015.9280395507812)
(6,1018.2131958007812)
(7,1020.4151611328125)
(8,1022.5458984375)
(9,1024.6143798828125)
(10,1026.6278076171875)
(11,1028.591552734375)
(12,1030.51025390625)
(13,1032.3875732421875)
(14,1034.22705078125)
(15,1036.0311279296875)
(16,1037.8023681640625)
(17,1039.5428466796875)
(18,1041.25439453125)
(19,1042.9385986328125)
(20,1044.596923828125)
(21,1046.230712890625)
(22,1047.8411865234375)
(23,1049.429443359375)
(24,1050.996337890625)
(25,1052.5428466796875)
(26,1054.06982421875)
(27,1055.5782470703125)
(28,1057.068359375)
(29,1058.541259765625)
(30,1059.997314453125)
(31,1061.4371337890625)
(32,1062.861328125)
(33,1064.2703857421875)
(34,1065.664794921875)
(35,1067.044921875)
(36,1068.4111328125)
(37,1069.76416015625)
(38,1071.10400390625)
(39,1072.43115234375)
(40,1073.7459716796875)
(41,1075.048828125)
(42,1076.33984375)
(43,1077.6195068359375)
(44,1078.887939453125)
(45,1080.1456298828125)
(46,1081.392578125)
(47,1082.629150390625)
(48,1083.85546875)
(49,1085.0718994140625)
(50,1086.278564453125)
(51,1087.475830078125)
(52,1088.6636962890625)
(53,1089.8424072265625)
(54,1091.01220703125)
(55,1092.173095703125)
(56,1093.3255615234375)
(57,1094.469482421875)
(58,1095.60498046875)
(59,1096.732421875)
(60,1097.8519287109375)
(61,1098.9635009765625)
(62,1100.0673828125)
(63,1101.16357421875)
(64,1102.25244140625)
(65,1103.3338623046875)
(66,1104.4080810546875)
(67,1105.475341796875)
(68,1106.535400390625)
(69,1107.5887451171875)
(70,1108.63525390625)
(71,1109.675048828125)
(72,1110.708251953125)
(73,1111.735107421875)
(74,1112.7554931640625)
(75,1113.76953125)
(76,1114.777587890625)
(77,1115.779296875)
(78,1116.775146484375)
(79,1117.7650146484375)
(80,1118.7490234375)
(81,1119.727294921875)
(82,1120.6998291015625)
(83,1121.666748046875)
(84,1122.6280517578125)
(85,1123.583984375)
(86,1124.534423828125)
(87,1125.4794921875)
(88,1126.41943359375)
(89,1127.35400390625)
(90,1128.283447265625)
(91,1129.2078857421875)
(92,1130.127197265625)
(93,1131.0416259765625)
(94,1131.951171875)
(95,1132.85595703125)
(96,1133.755615234375)
(97,1134.65087890625)
(98,1135.5413818359375)
(99,1136.42724609375)
(100,1137.3084716796875)
};
\addlegendentry{Closed-Form}

\draw[<->, thick, darkgreen, dashed] (axis cs:100,1031.6461255419356) -- (axis cs:100,1137.3084716796875);
\node at (87, 1084.4772986108114) {\textcolor{darkgreen}{$10.24\%$}};

\draw[<->, thick, purple, dashed] (axis cs:10,1008.1638648075632) -- (axis cs:10,1026.6278076171875);
\node at (20, 1020.8597790219997) {\textcolor{purple}{$1.80\%$}};

\end{axis}
\end{tikzpicture}
\caption{Comparison between the Sinkhorn-Knopp (i.e. Algorithm~\ref{alg:sinkhorn}) and the closed-form solution (i.e. Algorithm~\ref{alg:ot-closed-form}). The plot is made based on the data of ResNet20 trained on Cifar10. The relative difference is computed by (red - blue) / blue. } 
\label{fig:ot_objective}
\end{figure}
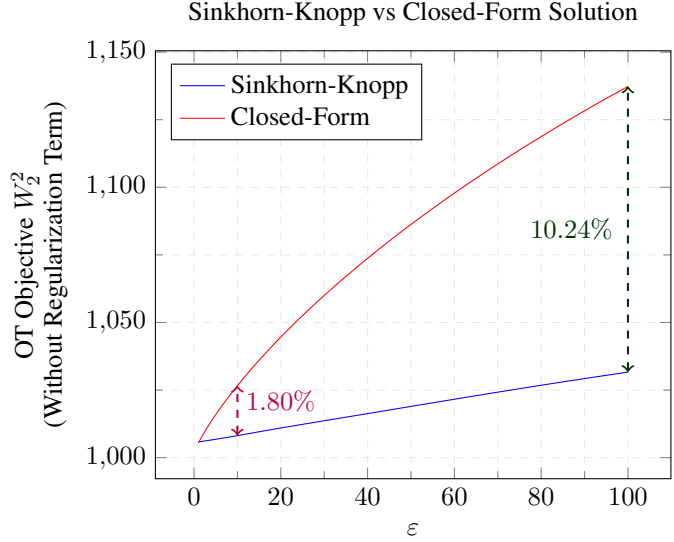

As depicted in \figurename~\ref{fig:ot_objective}, a comparative evaluation of the two algorithms in terms of attaining the \gls{ot} objective, $\ot$, is presented. The entropic regularization term's value has been omitted considering its non-impact on the pruning optimization of $\vec{w}$, given a constant $\vec{\Pi}$. Notably, the disparity between the two algorithms in their objective optimization magnifies as $\varepsilon$ increases. Typically, in real-world applications, the value of $\varepsilon$ oscillates between 0 and 10 (and we use $\varepsilon=1$ most frequently in our experiments). The variance in the performance of the two algorithms concerning \gls{ot} planning remains trivial, echoing our practical observations during the algorithmic implementation in this study.

\subsection{Derivative of $Q(\vec{w})$}
\label{sec:Q(w)}


Let's start by revisiting the function \(Q(\vec{w})\):
\begin{align}
Q(\vec{w}) &= \left\{\sum_{i=1}^n\sum_{j=1}^n\norm{x_i(\vec{w})-y_j}^2\pi_{ij}\right\} + \lambda\norm{\vec{w} - \vec{\bw}}^2 \label{eq:q_function}
\end{align}

Given:
\begin{align*}
x_i(\vec{w}) &= \nel_i^\top\vec{w} \\
y_j &= \nel_j^\top\vec{\bw}
\end{align*}

\textbf{Differentiating \(Q(\vec{w})\)} with respect to \(\vec{w}\):
\begin{align}
\nabla Q(\vec{w}) &= \nabla \left[ \left\{ \sum_{i=1}^n \sum_{j=1}^n \norm{\nel_i^\top\vec{w} - \nel_j^\top\vec{\bw}}^2 \pi_{ij} \right\} + \lambda \norm{\vec{w} - \vec{\bw}}^2 \right] \nonumber \\
&= 2\lambda(\vec{w} - \vec{\bw}) + \nabla \left[ \left\{ \sum_{i=1}^n \sum_{j=1}^n \norm{\nel_i^\top\vec{w} - \nel_j^\top\vec{\bw}}^2 \pi_{ij} \right\} \right] \label{eq:diff_q}
\end{align}

For the gradient of the inner term, consider:

\begin{align}
\nabla \left[ \left\{ \sum_{i=1}^n \sum_{j=1}^n \norm{\nel_i^\top\vec{w} - \nel_j^\top\vec{\bw}}^2 \pi_{ij} \right\} \right] &= \sum_{i=1}^n\sum_{j=1}^n 2\pi_{ij}(\nel_i^\top\vec{w} - \nel_j^\top\vec{\bw})\nel_i \label{eq:inner_diff}
\end{align}

\textbf{Expressing in Matrix Form}.
Given matrices:
\begin{align*}
\vec{G} &= \begin{bmatrix} {\nel}_1 \\ {\nel}_2 \\ \vdots \\ {\nel}_n \end{bmatrix} \\
\vec{\Pi} &= \text{Matrix with elements } \pi_{ij}
\end{align*}

Consider the double summation:
\[
\sum_{i=1}^n\sum_{j=1}^n 2\pi_{ij}(\nel_i^\top\vec{w} - \nel_j^\top\vec{\bw})\nel_i
\]

For each \( i \), the term \(\nel_i^\top\vec{w}\) projects vector \( \vec{w} \) onto \( \nel_i \). To compute this for every \( i \) simultaneously, it is 
$\vec{G} \vec{w}$. This results in an \( n \times 1 \) column vector. Similarly, for each \( j \), it is 
$\vec{G} \vec{\bw}$. This also produces an \( n \times 1 \) column vector.

The difference between these two projections for each \( i \) and \( j \) is \(
\vec{G} \vec{w} - \vec{G} \vec{\bw}
\). This results in an \( n \times 1 \) column vector. To incorporate the \( \pi_{ij} \) weights, we have
\[
\vec{\Pi} \left( \vec{G} \vec{w} - \vec{G} \vec{\bw} \right)
\]

This operation gives an \( n \times 1 \) column vector, where each element is a summation over \( j \) for the term \( \pi_{ij}(\nel_i^\top\vec{w} - \nel_j^\top\vec{\bw}) \).

To finalize the summation, we multiply it by $\vec{G}^\top$, yielding
\begin{equation}
\vec{G}^\top \left( \vec{\Pi} \left( \vec{G} \vec{w} - \vec{G} \vec{\bw} \right) \right).
\label{eq:matrix_form}
\end{equation}

Combining \eqrefp{eq:diff_q}, \eqrefp{eq:inner_diff}, and \eqrefp{eq:matrix_form}, we get:
\begin{align}
\nabla Q(\vec{w}) &= 2\lambda(\vec{w} - \vec{\bw}) + 2\vec{G}^\top (\vec{\Pi} (\vec{G}\vec{w} - \vec{G}\vec{\bw})) \nonumber \\
&= 2[\vec{G}^\top (\vec{\Pi} (\vec{G}\vec{w} - \vec{G}\vec{\bw})) + \lambda(\vec{w} - \vec{\bw})] \label{eq:final_form}
\end{align}

\subsection{Experiment Setup}
\label{sec:experiment_setup}

The models MLPNet, ResNet20, and MobileNetV1 underwent a pre-training phase of 100 epochs utilizing 4 NVIDIA Tesla V100 32 GB GPUs connected with NVlink. The training times were approximately 1 hour for MLPNet, 3 hours for ResNet20, and 1 day for MobileNetV1. Pre-pruning accuracy levels for these models are detailed under their respective names in \tablename~\ref{tab:accuracy}. For the pruning process, we either utilized 2 NVIDIA Tesla V100 32 GB GPUs with NVlink or a single Tesla A100 PCIE (available in 40 or 80 GB configurations). It's worth emphasizing the time-intensive nature of training and pruning the MobileNetV1 on ImageNet; thus, harnessing multiple GPUs for concurrent training is highly recommended.

In \tablename~\ref{tab:accuracy}, we set the pruning stage of LR and EWR to be $15$ for MLPNet and ResNet20 and $10$ for MobileNetV1. The sparsity $k_1, k_2, \ldots k_T$ in Algorithm~\ref{alg:cap} is arranged following an exponential gradual pruning schedule
\[
k_t = k_{T} + (k_{0}-k_{T})\left(1-\frac{t}{T}\right)^3
\]
with the initial sparsity $k_0$ set to zero. The fisher sample size setup follows~\citep[Table 2]{chen2022network}, shown as \tablename~\ref{tab:fisher-samples} of this paper below.
\begin{table}[H]
\centering
\caption{Comparison of Sample and Batch Sizes for Different Models}
\begin{tabular}{c|cc|cc|cc}
\toprule
Model & \multicolumn{2}{c|}{MLPNet} & \multicolumn{2}{c|}{ResNet20/50} & \multicolumn{2}{c}{MobileNet} \\
\cmidrule(lr){2-7}
 & Sample & Batch & Sample & Batch & Sample & Batch \\
\midrule
WF \& CBS & 1000 & 1 & 1000 & 1 & 400 & 2400 \\
LR \& EWR & 1000 & 1 & 1000 & 1 & 1000 & 16 \\
\bottomrule
\end{tabular}
\label{tab:fisher-samples}
\end{table}

In \tablename~\ref{tab:loss_resnet20}, \tablename~\ref{tab:loss_mobilenetv1}, \tablename~\ref{tab:accuracy_resnet20}, \tablename~\ref{tab:accuracy_mobilenetv1}, and additional results provided in the \appendixname, sparsity is set using a linear gradual pruning strategy, progressing from 0 to 0.75 or 0.95 across ten distinct stages for MLPNet and ResNet, and from 0 to 0.75 across eight distinct stages for MobileNetV1. The values are computed with linear incremental steps, from zero to the target sparsity. Notably, all recorded loss values are captured immediately post-pruning, devoid of any subsequent fine-tuning. This approach ensures that the loss values exclusively reflect the impact of the pruning algorithms, without being clouded by external factors. Contrasting with \tablename~\ref{tab:accuracy}, where each row represents a full pruning cycle, the loss values here are recorded across the ten incremental pruning stages. The empirical fisher is computed based on 100 samples with a batch size of 1. 

In calibrating noise for data in neural networks, we start with a well-trained network. First, we calculate the standard deviation $\sigma$ of the network's derivative. Then, we add Gaussian noise with zero mean to the data. After adding the noise, the standard deviation of the network's derivative changes to a new value, $\sigma'$, which is always greater than sigma. The goal is to adjust the standard deviation of the Gaussian noise so that $\sigma'$ becomes $\sigma'=\sigma+\sigma$ (referred to as noise level being $\sigma$) or $\sigma'=\sigma+2\sigma$  (referred to as noise level being $2\sigma$). 

Throughout the paper, we set $\lambda$ in the optimization problem~\eqrefp{eq:wasserstein_formulation_empirical} to $0.01$. The regularization multiplier $\varepsilon$ is set to $1$ unless specified otherwise. The noise level $\sigma$ is set to be the standard deviation of the original gradients.

\subsection{Ablation Study}
\label{sec:ablation_study}

\begin{table}[H]
\centering
\caption{Comparison of loss values in terms of $\varepsilon$ for ResNet20. The results are obtained from 25 runs, with 10\% Noisy data and noise level $\sigma$. The target sparsity is 0.95.}
\label{tab:my_label}
\begin{tabular}{c|c|*{5}{c}}
\toprule
Sparsity & LR Loss & \multicolumn{5}{c}{EWR Loss} \\
\cmidrule(lr){3-7}
         &         & \(\varepsilon = 0\) & \(\varepsilon = 1\) & \(\varepsilon = 2\) & \(\varepsilon = 10\) & \(\varepsilon = \infty\) \\
\midrule
0.95 & 2.89 & \textcolor{teal}{\textbf{2.77}} & \textcolor{teal}{\text{2.78}} & \textcolor{teal}{\text{2.78}} & \textcolor{teal}{\text{2.79}} & \textcolor{dtured}{\textbf{2.97}} \\
0.84 & 1.76 & \textcolor{teal}{\text{1.69}} & \textcolor{teal}{\textbf{1.61}} & \textcolor{teal}{\text{1.68}} & \textcolor{teal}{\text{1.71}} & \textcolor{dtured}{\textbf{2.62}} \\
0.74 & 0.68 & 0.68 & \textcolor{teal}{\text{0.67}} & \textcolor{teal}{\textbf{0.66}} & \textcolor{teal}{\textbf{0.66}} & \textcolor{dtured}{\textbf{0.81}} \\
0.63 & 0.36 & 0.36 & 0.36 & 0.36 & 0.36 &  0.37\\
0.53 & 0.31 & 0.31 & 0.31 & 0.31 & 0.31 &  0.31\\
0.42 & 0.31 & 0.31 & 0.31 & 0.31 & 0.31 &  0.31\\
0.32 & 0.30 & 0.30 & 0.30 & 0.30 & 0.30 &  0.30\\
0.21 & 0.30 & 0.30 & 0.30 & 0.30 & 0.30 &  0.30\\
0.11 & 0.31 & 0.31 & 0.31 & 0.31 & 0.31 &  0.31\\
\bottomrule
\end{tabular}
\end{table}
\begin{figure}[H]
    \centering
\begin{tikzpicture}
    \begin{axis}[
        width=0.7\textwidth,
        height=0.35\textwidth,
        ybar, 
        enlargelimits=0.15,
        legend style={at={(0.5,-0.2)}, anchor=north,legend columns=-1},
        legend cell align={left},
        ylabel style={align=center},
        ylabel=Loss Reduction \\ (LR - EWR),
        symbolic x coords={$\varepsilon = 0$, $\varepsilon = 1$, $\varepsilon = 2$, $\varepsilon = 10$},
        grid=both,
        minor tick num=1,
        grid style={line width=.1pt, dashed, draw=black!10},
        xtick=data,
        yticklabel style={
            /pgf/number format/precision=2,
            /pgf/number format/fixed},
        error bars/.cd,
        ]
        \addplot+[error bars/.cd, y dir=both, y explicit] coordinates {
            ($\varepsilon = 0$, 0.0) 
            ($\varepsilon = 1$, 0.01) +- (0.0075,0.0075)
            ($\varepsilon = 2$, 0.02) +- (0.007,0.007)
            ($\varepsilon = 10$, 0.02) +- (0.008,0.008)
        }; 
        \addplot+[error bars/.cd, y dir=both, y explicit] coordinates {
            ($\varepsilon = 0$, 0.07) +- (0.0113,0.0123)
            ($\varepsilon = 1$, 0.15) +- (0.0123,0.0123)
            ($\varepsilon = 2$, 0.08) +- (0.0131,0.0131)
            ($\varepsilon = 10$, 0.05) +- (0.0139,0.0139)
        }; 
        \addplot+[error bars/.cd, y dir=both, y explicit] coordinates {
            ($\varepsilon = 0$, 0.12) +- (0.0121,0.0121)
            ($\varepsilon = 1$, 0.11) +- (0.0135,0.0135)
            ($\varepsilon = 2$, 0.11) +- (0.0133,0.0133)
            ($\varepsilon = 10$, 0.10) +- (0.0117,0.0117)
        }; 
        \legend{Sparsity = 0.74~~~~~, Sparsity = 0.84~~~~~, Sparsity = 0.95}
    \end{axis}
\end{tikzpicture}
    \caption{Loss reduction with different $\varepsilon$. The result is averaged over 25 runs for ResNet20, with 10\% Noisy data and noise level $\sigma$. The error bar shows 90\% confidence interval. The target sparsity is 0.95.}
    \label{fig:enter-label2}
\end{figure}
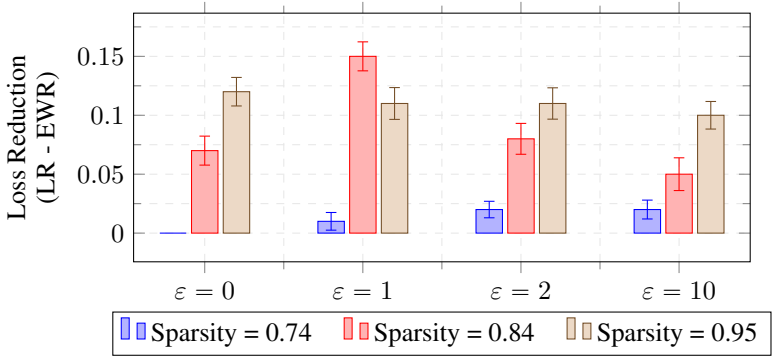

The ablation study centered on ResNet20 gives insight into the influence of the entropic regularization multiplier, $\varepsilon$, on pruning. The aim is to understand how different values of this parameter affect the loss $\L(\vec{w})$ during the pruning process. 

\textbf{Observations from Loss Values}.
From Table~\ref{tab:my_label}, we can glean the following points. For higher sparsity levels (0.95, 0.84, 0.74), EWR Loss consistently outperforms the LR Loss, except for $\varepsilon=\infty$. The optimum performance of the EWR Loss, across varying sparsities, tends to occur when $\varepsilon$ is set at values between 1 and 2. As $\varepsilon$ tends towards infinity, the EWR loss exceeds the LR loss. This phenomenon aligns with the neighborhood interpolation elaborated upon in Section~\ref{sec:theoretical}. Specifically, incorporating an excessive number of distant data points into the interpolation detrimentally impacts performance. For lower sparsity levels (from 0.63 downwards), the differences between LR Loss and EWR Loss across different $\varepsilon$ values are minuscule.

\textbf{Loss Reduction Insights}.
Referencing Figure~\ref{fig:enter-label2}, it's evident that the loss reduction (difference between LR and EWR) is more pronounced at higher sparsity levels. For a sparsity of 0.84, $\varepsilon = 1$ demonstrates the most significant loss reduction.
 The trend of EWR loss for different $\varepsilon$ values is consistent across varying sparsity levels. The impact $\varepsilon$ is clearly visible at higher sparsity levels. For mid to high sparsity levels, lower values of $\varepsilon$ (specifically 1 and 2) seem to achieve the best balance in terms of loss.

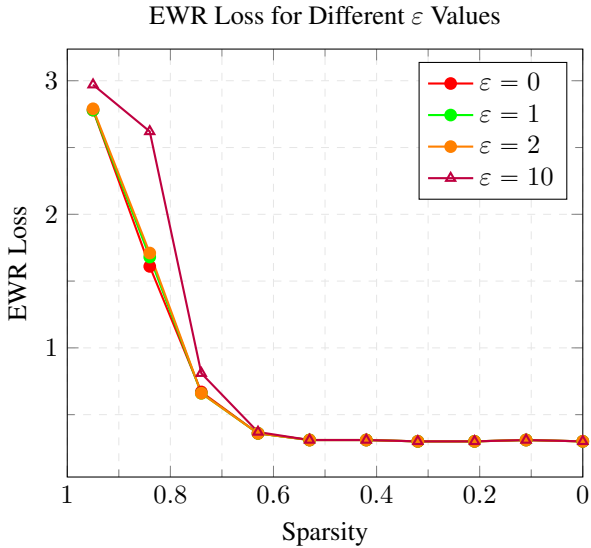
\begin{figure}[H]
    \centering
\begin{tikzpicture}
\begin{axis}[
  title={EWR Loss for Different $\varepsilon$ Values},
  xlabel={Sparsity},
  ylabel={EWR Loss},
  legend pos={north east},
  legend cell align={left},
grid=both,
minor tick num=1,
grid style={line width=.1pt, dashed, draw=black!10},
  xmin=0, xmax=1,
  x dir=reverse
]

\addlegendentry{$\varepsilon = 0$}

\addplot[red, mark=*, thick] coordinates {
  (0.95, 2.78)
  (0.84, 1.61)
  (0.74, 0.67)
  (0.63, 0.36)
  (0.53, 0.31)
  (0.42, 0.31)
  (0.32, 0.30)
  (0.21, 0.30)
  (0.11, 0.31)
  (0.00, 0.30)
};
\addlegendentry{$\varepsilon = 1$}

\addplot[green, mark=*, thick] coordinates {
  (0.95, 2.78)
  (0.84, 1.68)
  (0.74, 0.66)
  (0.63, 0.36)
  (0.53, 0.31)
  (0.42, 0.31)
  (0.32, 0.30)
  (0.21, 0.30)
  (0.11, 0.31)
  (0.00, 0.30)
};
\addlegendentry{$\varepsilon = 2$}

\addplot[orange, mark=*, thick] coordinates {
  (0.95, 2.79)
  (0.84, 1.71)
  (0.74, 0.66)
  (0.63, 0.36)
  (0.53, 0.31)
  (0.42, 0.31)
  (0.32, 0.30)
  (0.21, 0.30)
  (0.11, 0.31)
  (0.00, 0.30)
};
\addlegendentry{$\varepsilon = 10$}

\addplot[purple, mark=triangle , thick] coordinates {
  (0.95, 2.97)
  (0.84, 2.62)
  (0.74, 0.81)
  (0.63, 0.37)
  (0.53, 0.31)
  (0.42, 0.31)
  (0.32, 0.30)
  (0.21, 0.30)
  (0.11, 0.31)
  (0.00, 0.30)
};
\addlegendentry{$\varepsilon = \infty$}
\end{axis}
\end{tikzpicture}
\caption{EWR Loss in function of the sparsity. The result is obtained over 25 runs on ResNet20, with 10\% Noisy data and noise level $\sigma$. The target sparsity is 0.95. }
\label{fig:EWR_loss}
\end{figure}

\figurename~\ref{fig:EWR_loss} illustrates EWR Loss for various $ \varepsilon $ values against sparsity, there is a clear trend of decreasing loss as sparsity decreases, consistent across all $ \varepsilon $ values. Particularly at $ \varepsilon = \infty $, there's a pronounced increase in loss at higher sparsity, suggesting that extreme entropic regularization might hinder optimal pruning. However, at low sparsity levels, the loss is consistent across all $ \varepsilon $, emphasizing the minimal impact of $ \varepsilon $ in limited pruning scenarios. This underscores the significance of choosing an appropriate $ \varepsilon $, balancing between regularization and pruning efficiency.

The ablation study provides insights into the role of the parameter \( \varepsilon \) in pruning. Its influence diminishes at low sparsity levels but becomes significant at extremely high sparsity. For ResNet20, an optimal range for \( \varepsilon \) appears to be between 1 and 2, ensuring effective pruning. Generally, while the exact choice of \( \varepsilon \) doesn't drastically alter the pruning outcome, exceedingly high values, such as \(1\times 10^8\), might lead to less than ideal pruning decisions in very sparse networks.

\subsection{Analysis of Pruning Stage Versus Performance}

\textbf{Analysis of Loss}. \figurename~\ref{fig:loss_diff_resnet20} depicts the relationship between the sparsity and the difference in loss (LR - EWR) under the influence of varying noise levels.

\begin{figure}[t]
\centering
\begin{minipage}{0.49\textwidth}
\begin{tikzpicture}
\begin{axis}[
    width=\textwidth, 
    height=0.35\textheight,
    title = {10\% Noisy Data},
    xlabel = {Sparsity},
    ylabel = {Difference in Loss (LR - EWR)},
    grid=both,
    minor tick num=1,
    grid style={line width=.1pt, dashed, draw=black!10},
    legend pos = {north west},
    legend cell align={left},
    yticklabel style={
    /pgf/number format/precision=2,
    /pgf/number format/fixed},
]
  \addplot[purple, thick, solid, mark=*, mark options={fill=purple}] coordinates {
    (0.95, 0.121001) 
    (0.84, 0.080652) 
    (0.74, 0.060929) 
    (0.63, 0.008523)
    (0.53, 0.000420)
    (0.42, 0.000064)
    (0.32, 0.000018)
    (0.21, 0.000057)
    (0.10, 0.000008)
    (0.00, 0)
  };
    \addplot[cyan, thick, solid, mark=*, mark options={fill=cyan}] coordinates {
    (0.95, 0.081239) 
    (0.84, 0.043125) 
    (0.74, 0.020007) 
    (0.63, 0.000201)
    (0.53, 0.000330)
    (0.42, -0.000035)
    (0.32, -0.000326)
    (0.21, -0.000062)
    (0.10, 0.000005)
    (0.00, 0)
  };
  \legend{Noise = $2\sigma$, Noise = $\sigma$}

  \addplot+[purple, error bars/.cd, y dir=both, y explicit] coordinates {
    (0.95, 0.121001) +- (0, 0.015)
    (0.84, 0.080652) +- (0, 0.014)
    (0.74, 0.060929) +- (0, 0.012)
    (0.63, 0.008523) +- (0, 0.007)
  };

  \node[above, text=purple] at (axis cs:0.89, 0.121001) {\footnotesize 4.35\%};
  \node[above, text=purple] at (axis cs:0.77, 0.080652) {\footnotesize 5.15\%};
  \node[above, text=purple] at (axis cs:0.67, 0.060929) {\footnotesize 1.32\%};
  \node[above, text=purple] at (axis cs:0.57, 0.008523) {\footnotesize 0.85\%};

  \addplot+[cyan, error bars/.cd, y dir=both, y explicit] coordinates {
    (0.95, 0.081239) +- (0, 0.0175)
    (0.84, 0.043125) +- (0, 0.0113)
    (0.74, 0.020007) +- (0, 0.0119)
  };
  \node[above, text=cyan] at (axis cs:0.97, 0.05) {\footnotesize 2.87\%};
  \node[above, text=cyan] at (axis cs:0.86, 0.02) {\footnotesize 2.73\%};
  \node[above, text=cyan] at (axis cs:0.76, 0.0) {\footnotesize 3.03\%};
  
  \addplot[black, dashed] coordinates {
    (0.95, 0)
    (0.00, 0)
  };
\end{axis}
\end{tikzpicture}
\end{minipage}
\begin{minipage}{0.49\textwidth}
\begin{tikzpicture}
\begin{axis}[
    width=\textwidth, 
    height=0.35\textheight,
    title = {25\% Noisy Data},
    xlabel = {Sparsity},
    grid=both,
    minor tick num=1,
    grid style={line width=.1pt, dashed, draw=black!10},
    legend pos = {north west},
    legend cell align={left},
    yticklabel style={
    /pgf/number format/precision=2,
    /pgf/number format/fixed},
]
  \addplot[purple, thick, solid, mark=*, mark options={fill=purple}] coordinates {
    (0.95, 0.110487) 
    (0.84, 0.071355) 
    (0.74, 0.010546) 
    (0.63, 0.000679)
    (0.53, 0.000064)
    (0.42, -0.000081)
    (0.32, -0.000117)
    (0.21, -0.000022)
    (0.10, 0.000032)
    (0.00, 0)
  };
    \addplot[cyan, thick, solid, mark=*, mark options={fill=cyan}] coordinates {
    (0.95, 0.100481) 
    (0.84, 0.061355) 
    (0.74, 0.003337) 
    (0.63, 0.000320)
    (0.53, -0.000135)
    (0.42, -0.000108)
    (0.32, -0.000110)
    (0.21, -0.000033)
    (0.10, -0.000024)
    (0.00, 0)
  };
  \legend{Noise = $2\sigma$, Noise = $\sigma$}

Confidence intervals (error bars) for the three specified points
  \addplot+[purple, error bars/.cd, y dir=both, y explicit] coordinates {
    (0.95, 0.110487) +- (0, 0.015)
    (0.84, 0.071355) +- (0, 0.013)
    (0.74, 0.010546) +- (0, 0.013)
  };

  \node[above, text=purple] at (axis cs:0.87, 0.110487) {\footnotesize 3.82\%};
  \node[above, text=purple] at (axis cs:0.75, 0.071355) {\footnotesize 4.05\%};
  \node[above, text=purple] at (axis cs:0.65, 0.010546) {\footnotesize 1.55\%};

  \addplot+[cyan, error bars/.cd, y dir=both, y explicit] coordinates {
    (0.95, 0.100481) +- (0, 0.0135)
    (0.84, 0.061355) +- (0, 0.0123)
    (0.74, 0.003337) +- (0, 0.0074)
  };
  \node[above, text=cyan] at (axis cs:0.85, 0.094) {\footnotesize 3.50\%};
  \node[above, text=cyan] at (axis cs:0.94, 0.049) {\footnotesize 4.07\%};
  \node[above, text=cyan] at (axis cs:0.86, 0.005) {\footnotesize 0.49\%};
  
  \addplot[black, dashed] coordinates {
    (0.95, 0)
    (0.00, 0)
  };
\end{axis}
\end{tikzpicture}
\end{minipage}
\caption{Difference in loss between LR and EWR for ResNet20. The data is from \tablename~\ref{tab:loss_resnet20}. The relative loss improvement of EWR over LR is reported. The target sparsity is 0.95.}
\label{fig:loss_diff_resnet20}
\end{figure}
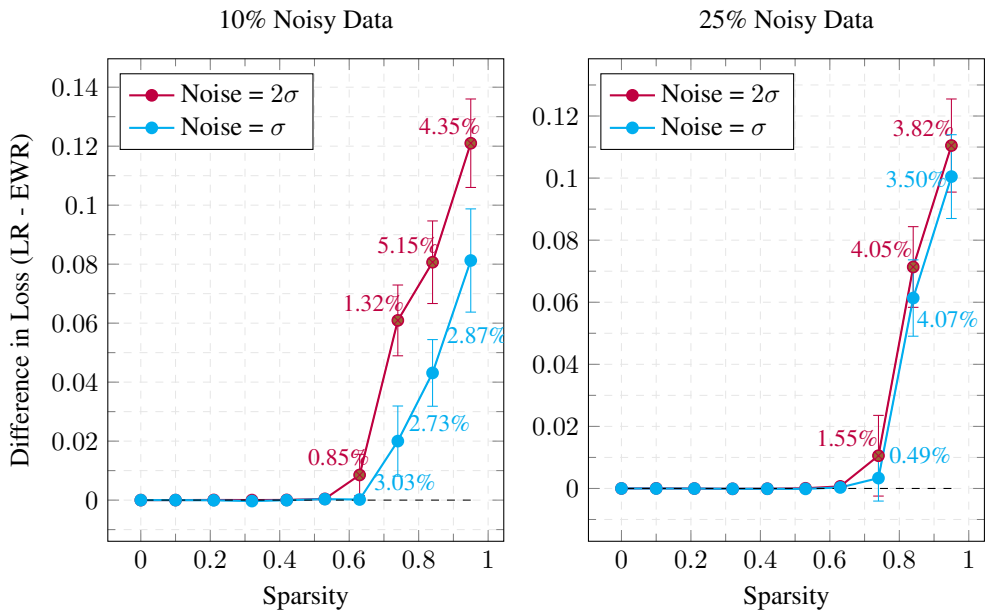
\begin{figure}[!h]
\centering
\begin{minipage}{0.49\textwidth}
\begin{tikzpicture}
\begin{axis}[
    width=\textwidth, 
    height=0.35\textheight,
    title = {10\% Noisy Data},
    xlabel = {Sparsity},
    ylabel = {Difference in Loss (LR - EWR)},
    grid=both,
    minor tick num=1,
    grid style={line width=.1pt, dashed, draw=black!10},
    legend pos = {north west},
    legend cell align={left},
    yticklabel style={
    /pgf/number format/precision=2,
    /pgf/number format/fixed},
]
  \addplot[purple, thick, solid, mark=*, mark options={fill=purple}] coordinates {
    (0.74, 0.220415) 
    (0.63, 0.080132)
    (0.53, 0.100001)
    (0.42, 0.000035)
    (0.32, 0.000021)
    (0.21, 0.000020)
    (0.10, 0.000007)
    (0.00, 0)
  };
    \addplot[cyan, thick, solid, mark=*, mark options={fill=cyan}] coordinates {
    (0.74, 0.200125) 
    (0.63, 0.070019)
    (0.53, 0.079923)
    (0.42, 0.000064)
    (0.32, 0.000018)
    (0.21, 0.000037)
    (0.10, 0.000006)
    (0.00, 0)
  };
  \legend{Noise = $2\sigma$, Noise = $\sigma$}

  \addplot+[purple, error bars/.cd, y dir=both, y explicit] coordinates {
    (0.74, 0.220415) +- (0, 0.024)
    (0.63, 0.080132) +- (0, 0.020)
    (0.53, 0.100001) +- (0, 0.014)
  };

  \node[above, text=purple] at (axis cs:0.64, 0.210125) {\footnotesize 5.02\%};
  \node[above, text=purple] at (axis cs:0.73, 0.0750132) {\footnotesize 3.23\%};
  \node[above, text=purple] at (axis cs:0.43, 0.089923) {\footnotesize 6.01\%};

  \addplot+[cyan, error bars/.cd, y dir=both, y explicit] coordinates {
    (0.74, 0.200125) +- (0, 0.021)
    (0.63, 0.070019) +- (0, 0.018)
    (0.53, 0.079923) +- (0, 0.013)
  };
  \node[above, text=cyan] at (axis cs:0.64, 0.190125) {\footnotesize 4.41\%};
  \node[above, text=cyan] at (axis cs:0.73, 0.060019) {\footnotesize 2.61\%};
  \node[above, text=cyan] at (axis cs:0.43, 0.069923) {\footnotesize 5.16\%};
  
  \addplot[black, dashed] coordinates {
    (0.74, 0)
    (0.00, 0)
  };
\end{axis}
\end{tikzpicture}
\end{minipage}
\begin{minipage}{0.49\textwidth}
\begin{tikzpicture}
\begin{axis}[
    width=\textwidth, 
    height=0.35\textheight,
    title = {25\% Noisy Data},
    xlabel = {Sparsity},
    grid=both,
    minor tick num=1,
    grid style={line width=.1pt, dashed, draw=black!10},
    legend pos = {north west},
    legend cell align={left},
    yticklabel style={
    /pgf/number format/precision=2,
    /pgf/number format/fixed},
]
  \addplot[purple, thick, solid, mark=*, mark options={fill=purple}] coordinates {
    (0.74, 0.359995) 
    (0.63, 0.120002)
    (0.53, 0.100301)
    (0.42, 0.000035)
    (0.32, 0.000021)
    (0.21, 0.000020)
    (0.10, 0.000007)
    (0.00, 0)
  };
    \addplot[cyan, thick, solid, mark=*, mark options={fill=cyan}] coordinates {
    (0.74, 0.400021) 
    (0.63, 0.102351)
    (0.53, 0.089911)
    (0.42, 0.000054)
    (0.32, 0.000028)
    (0.21, 0.000037)
    (0.10, 0.000005)
    (0.00, 0)
  };
  \legend{Noise = $2\sigma$, Noise = $\sigma$}

Confidence intervals (error bars) for the three specified points
  \addplot+[purple, error bars/.cd, y dir=both, y explicit] coordinates {
    (0.74, 0.359995) +- (0, 0.015)
    (0.63, 0.120002) +- (0, 0.013)
    (0.53, 0.100301) +- (0, 0.013)
  };

  \node[above, text=purple] at (axis cs:0.63, 0.339995) {\footnotesize 7.92\%};
  \node[above, text=purple] at (axis cs:0.73, 0.111355) {\footnotesize 5.00\%};
  \node[above, text=purple] at (axis cs:0.43, 0.09) {\footnotesize 6.63\%};

  \addplot+[cyan, error bars/.cd, y dir=both, y explicit] coordinates {
    (0.74, 0.400021) +- (0, 0.0135)
    (0.63, 0.102351) +- (0, 0.0123)
    (0.53, 0.089911) +- (0, 0.0074)
  };
  \node[above, text=cyan] at (axis cs:0.63, 0.379995) {\footnotesize 8.13\%};
  \node[above, text=cyan] at (axis cs:0.73, 0.082315) {\footnotesize 4.01\%};
  \node[above, text=cyan] at (axis cs:0.43, 0.065) {\footnotesize 5.16\%};
  
  \addplot[black, dashed] coordinates {
    (0.74, 0)
    (0.00, 0)
  };
\end{axis}
\end{tikzpicture}
\end{minipage}

\caption{Difference in loss between LR and EWR for MobileNetV1. The data is from \tablename~\ref{tab:loss_mobilenetv1}. The relative loss improvement of EWR over LR is reported. The target sparsity is 0.75.}
\label{fig:loss_diff_mobilenetv1}
\end{figure}
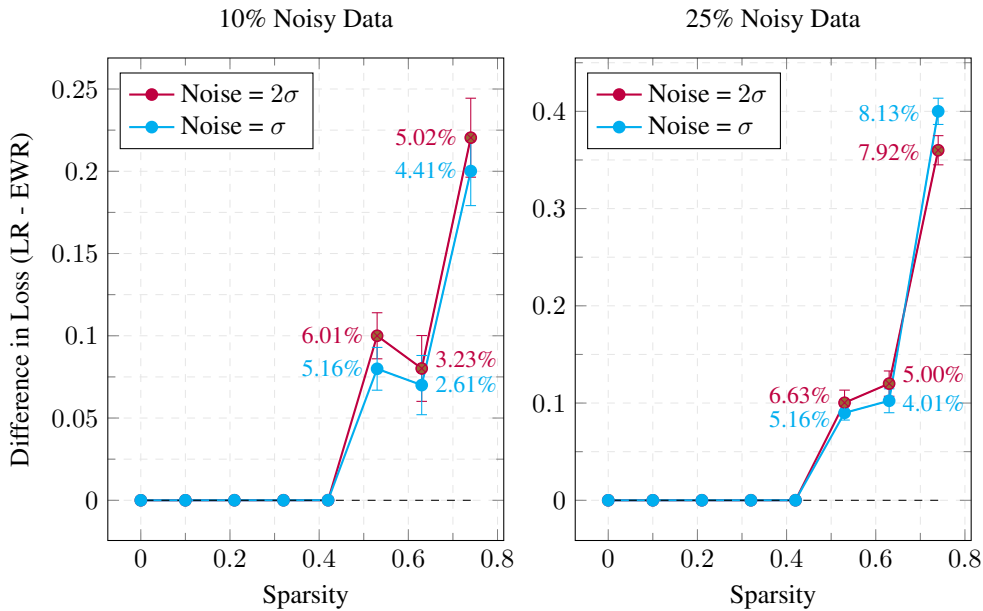

\begin{figure}[t]
    \centering
\begin{tikzpicture}
\begin{axis}[
    title={Pruning Stage vs. Accuracy (\%)},
    ylabel={Accuracy Difference (\%)},
    xlabel={Pruning Stage},
    ymajorgrids=true,
    xtick={1,2,3,4,5,6,7,8,9},
    boxplot/draw direction = y,
    enlarge x limits=0.05,
    yticklabel style={
        /pgf/number format/precision=2,
        /pgf/number format/fixed},
    grid=both,
    minor tick num=1,
    grid style={line width=.1pt, dashed, draw=black!10},
]
\addplot+[boxplot prepared={
    median=0.001999999999989,
    upper quartile=0.0020000000000072,
    lower quartile=0.0,
    upper whisker=0.0039999999999963,
    lower whisker=0.0,
}, fill, fill opacity=0.5, draw=black, solid] coordinates {};
\addplot+[boxplot prepared={
    median=0.0539999999999963,
    upper quartile=0.06,
    lower quartile=0.0220000000000072,
    upper whisker=0.0839999999999963,
    lower whisker=0.011999999999989,
}, fill, fill opacity=0.5, draw=black, solid] coordinates {};
\addplot+[boxplot prepared={
    median=0.03,
    upper quartile=0.0479999999999927,
    lower quartile=0.0160000000000036,
    upper whisker=0.0679999999999927,
    lower whisker=0.0079999999999927,
}, fill, fill opacity=0.5, draw=black, solid] coordinates {};
\addplot+[boxplot prepared={
    median=0.0279999999999927,
    upper quartile=0.04,
    lower quartile=0.0120000000000072,
    upper whisker=0.0539999999999963,
    lower whisker=0.0020000000000072,
}, fill, fill opacity=0.5, draw=black, solid] coordinates {};
\addplot+[boxplot prepared={
    median=0.0279999999999927,
    upper quartile=0.04,
    lower quartile=0.0220000000000072,
    upper whisker=0.0640000000000145,
    lower whisker=0.0039999999999963,
}, fill, fill opacity=0.5, draw=black, solid] coordinates {};
\addplot+[boxplot prepared={
    median=0.19,
    upper quartile=0.2020000000000072,
    lower quartile=0.1579999999999927,
    upper whisker=0.2479999999999927,
    lower whisker=0.0939999999999963,
}, fill, fill opacity=0.5, draw=black, solid] coordinates {};
\addplot+[boxplot prepared={
    median=0.7760000000000037,
    upper quartile=1.371999999999998,
    lower quartile=0.4140000000000055,
    upper whisker=1.7359999999999944,
    lower whisker=0.2020000000000072,
}, fill, fill opacity=0.5, draw=black, solid] coordinates {};
\addplot+[boxplot prepared={
    median=2.0319999999999983,
    upper quartile=2.498000000000002,
    lower quartile=1.6539999999999964,
    upper whisker=3.2260000000000035,
    lower whisker=1.1459999999999945,
}, fill, fill opacity=0.5, draw=black, solid] coordinates {};
\addplot+[boxplot prepared={
    median=0.3420000000000004,
    upper quartile=0.8120000000000005,
    lower quartile=0.2280000000000018,
    upper whisker=1.3579999999999996,
    lower whisker=0.0520000000000004,
}, fill, fill opacity=0.5, draw=black, solid] coordinates {};
\end{axis}
\end{tikzpicture}
~\\~\\
\begin{tikzpicture}
\begin{axis}[
    title={Pruning Stage vs. Top 5 Accuracy (\%)},
    ylabel={Top 5 Accuracy Difference (\%)},
    xlabel={Pruning Stage},
    ymajorgrids=true,
    xtick={1,2,3,4,5,6,7,8,9},
    boxplot/draw direction = y,
    enlarge x limits=0.05,
    yticklabel style={
        /pgf/number format/precision=2,
        /pgf/number format/fixed},
    grid=both,
    minor tick num=1,
    grid style={line width=.1pt, dashed, draw=black!10},
]
\addplot+[boxplot prepared={
    median=0.0,
    upper quartile=0.0,
    lower quartile=0.0,
    upper whisker=0.0,
    lower whisker=0.0,
}, fill, fill opacity=0.5, draw=black, solid] coordinates {};
\addplot+[boxplot prepared={
    median=0.0099999999999909,
    upper quartile=0.0139999999999957,
    lower quartile=0.0079999999999813,
    upper whisker=0.0160000000000053,
    lower whisker=0.0,
}, fill, fill opacity=0.5, draw=black, solid] coordinates {};
\addplot+[boxplot prepared={
    median=0.0099999999999909,
    upper quartile=0.0120000000000004,
    lower quartile=0.0079999999999955,
    upper whisker=0.01400000000001,
    lower whisker=0.0019999999999953,
}, fill, fill opacity=0.5, draw=black, solid] coordinates {};
\addplot+[boxplot prepared={
    median=0.0160000000000053,
    upper quartile=0.0180000000000148,
    lower quartile=0.0079999999999955,
    upper whisker=0.0259999999999962,
    lower whisker=0.0019999999999953,
}, fill, fill opacity=0.5, draw=black, solid] coordinates {};
\addplot+[boxplot prepared={
    median=0.0120000000000004,
    upper quartile=0.0180000000000006,
    lower quartile=0.0100000000000051,
    upper whisker=0.0220000000000055,
    lower whisker=0.0019999999999953,
}, fill, fill opacity=0.5, draw=black, solid] coordinates {};
\addplot+[boxplot prepared={
    median=0.0100000000000051,
    upper quartile=0.0220000000000055,
    lower quartile=0.0060000000000144,
    upper whisker=0.034000000000006,
    lower whisker=0.0019999999999953,
}, fill, fill opacity=0.5, draw=black, solid] coordinates {};
\addplot+[boxplot prepared={
    median=0.169999999999989,
    upper quartile=0.200000000000043,
    lower quartile=0.130000000000008,
    upper whisker=0.340000000000003,
    lower whisker=0.060000000000025,
}, fill, fill opacity=0.5, draw=black, solid] coordinates {};
\addplot+[boxplot prepared={
    median=1.3259999999999962,
    upper quartile=1.7060000000000065,
    lower quartile=0.9019999999999981,
    upper whisker=2.2639999999999958,
    lower whisker=0.5200000000000102,
}, fill, fill opacity=0.5, draw=black, solid] coordinates {};
\addplot+[boxplot prepared={
    median=1.5659999999999954,
    upper quartile=1.9080000000000084,
    lower quartile=1.1600000000000051,
    upper whisker=3.054000000000009,
    lower whisker=0.7759999999999976,
}, fill, fill opacity=0.5, draw=black, solid] coordinates {};
\end{axis}
\end{tikzpicture}
    \caption{Accuracy difference vs. pruning stage for ResNet20. The difference is defined to be (EWR - LR) / LR. The data is obtained over 5-25 runs over combinations of different setups: noise level $\sigma$ and $2\sigma$, noisy gradient proportion 10\% and 25\%, $\varepsilon=1,2$. The target sparsity is 0.95.}
    \label{fig:accuracy_resnet20}
\end{figure}
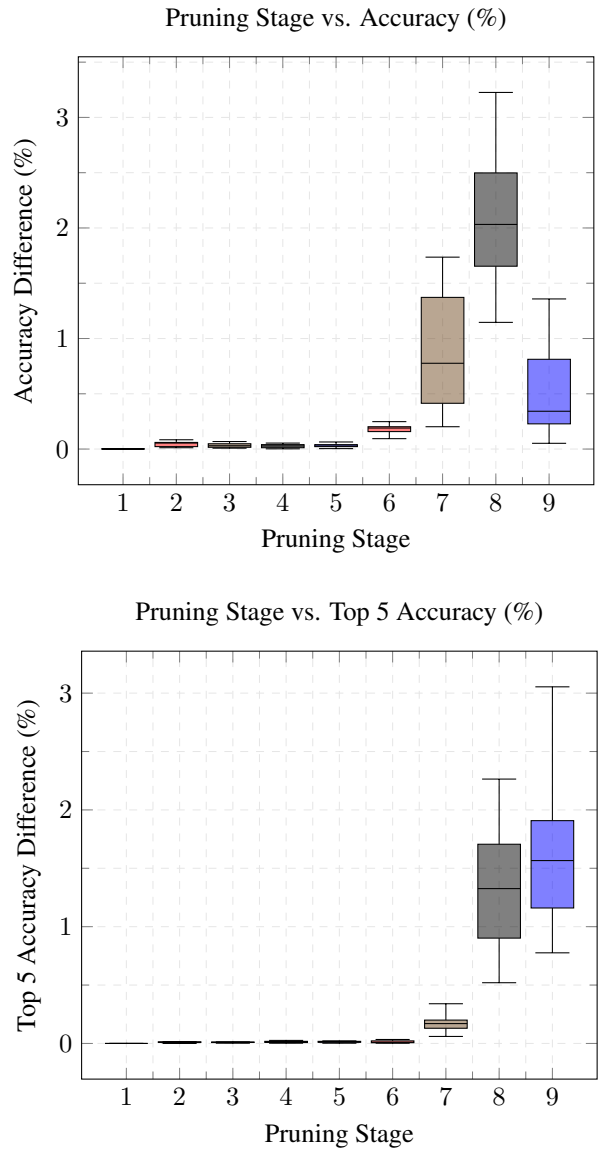

\begin{figure}[t]
    \centering
\begin{tikzpicture}
\begin{axis}[
    title={Pruning Stage vs. Accuracy (\%)},
    ylabel={Accuracy Difference (\%)},
    xlabel={Pruning Stage},
    ymajorgrids=true,
    xtick={1,2,3,4,5,6,7},
    boxplot/draw direction = y,
    enlarge x limits=0.05,
    yticklabel style={
        /pgf/number format/precision=2,
        /pgf/number format/fixed},
    grid=both,
    minor tick num=1,
    grid style={line width=.1pt, dashed, draw=black!10},
]
\addplot+[boxplot prepared={
    median=1.462939705144245e-06,
    upper quartile=0.01808562724325409,
    lower quartile=-0.02469958095996934,
    upper whisker=0.05127124596151145,
    lower whisker=-0.07773583846660331,
}, fill, fill opacity=0.5, draw=black, solid] coordinates {};
\addplot+[boxplot prepared={
    median=0.019295530730706965,
    upper quartile=0.05616708012290648,
    lower quartile=-0.022676433231229982,
    upper whisker=0.06451884765030733,
    lower whisker=-0.04632944446784316,
}, fill, fill opacity=0.5, draw=black, solid] coordinates {};
\addplot+[boxplot prepared={
    median=0.10931839147920547,
    upper quartile=0.1341656390624168,
    lower quartile=0.08290478755097684,
    upper whisker=0.14672285839966254,
    lower whisker=0.06564849917867914,
}, fill, fill opacity=0.5, draw=black, solid] coordinates {};
\addplot+[boxplot prepared={
    median=0.15462062447755315,
    upper quartile=0.17393977502197355,
    lower quartile=0.10568238576685965,
    upper whisker=0.22454223023036027,
    lower whisker=0.15216895900259497,
}, fill, fill opacity=0.5, draw=black, solid] coordinates {};
\addplot+[boxplot prepared={
    median=1.8742356240737736,
    upper quartile=2.6245968514094784,
    lower quartile=1.0691340149770492,
    upper whisker=3.1866827168840595,
    lower whisker=0.3428270042194093,
}, fill, fill opacity=0.5, draw=black, solid] coordinates {};
\addplot+[boxplot prepared={
    median=1.646770434981006,
    upper quartile=2.2207526779014164,
    lower quartile=0.26834327079226106,
    upper whisker=3.016623353347074,
    lower whisker=1.3380784172091476,
}, fill, fill opacity=0.5, draw=black, solid] coordinates {};
\addplot+[boxplot prepared={
    median=6.005255204374878,
    upper quartile=12.620691825323785,
    lower quartile=4.0033887098565037466,
    upper whisker=16.285785198125186,
    lower whisker=2.848103962505326,
}, fill, fill opacity=0.5, draw=black, solid] coordinates {};

\end{axis}
\end{tikzpicture}
~\\~\\
\begin{tikzpicture}
\begin{axis}[
    title={Pruning Stage vs. Top 5 Accuracy (\%)},
    ylabel={Top 5 Accuracy Difference (\%)},
    xlabel={Pruning Stage},
    ymajorgrids=true,
    xtick={1,2,3,4,5,6,7},
    boxplot/draw direction = y,
    enlarge x limits=0.05,
    yticklabel style={
        /pgf/number format/precision=2,
        /pgf/number format/fixed},
    grid=both,
    minor tick num=1,
    grid style={line width=.1pt, dashed, draw=black!10},
]
\addplot+[boxplot prepared={
    median=0.013500329718986182,
    upper quartile=0.026630371488419104,
    lower quartile=0.0008336084333439731,
    upper whisker=0.04549287648130468,
    lower whisker=-0.01663893510816947,
}, fill, fill opacity=0.5, draw=black, solid] coordinates {};
\addplot+[boxplot prepared={
    median=-0.01534947035860815,
    upper quartile=-0.007487803274868848,
    lower quartile=-0.02663479646146009,
    upper whisker=0.011098779134300907,
    lower whisker=-0.0170156099726242,
}, fill, fill opacity=0.5, draw=black, solid] coordinates {};
\addplot+[boxplot prepared={
    median=0.028726109587345602,
    upper quartile=0.037373794049671116,
    lower quartile=0.0181325410942678,
    upper whisker=0.05245652804749776,
    lower whisker=-0.002787844995815697,
}, fill, fill opacity=0.5, draw=black, solid] coordinates {};
\addplot+[boxplot prepared={
    median=-0.016603953847047333,
    upper quartile=0.006938618769853816,
    lower quartile=-0.05879013740526883,
    upper whisker=0.008492328596493438,
    lower whisker=-0.1162746800558695,
}, fill, fill opacity=0.5, draw=black, solid] coordinates {};
\addplot+[boxplot prepared={
    median=1.0958641550719064,
    upper quartile=1.4822922181353748,
    lower quartile=0.6379614830309782,
    upper whisker=1.6316876740217325,
    lower whisker=0.2741422002122411,
}, fill, fill opacity=0.5, draw=black, solid] coordinates {};
\addplot+[boxplot prepared={
    median=0.7925302387997408,
    upper quartile=1.1886377084253708,
    lower quartile=0.12097648180594,
    upper whisker=2.063302687907922,
    lower whisker=0.6879777623349612,
}, fill, fill opacity=0.5, draw=black, solid] coordinates {};
\addplot+[boxplot prepared={
    median=5.413705964555613,
    upper quartile=9.352955290389403,
    lower quartile=3.053911919896228,
    upper whisker=10.185463659147858,
    lower whisker=2.040230605339013,
}, fill, fill opacity=0.5, draw=black, solid] coordinates {};

\end{axis}
\end{tikzpicture}
    \caption{Accuracy difference vs. pruning stage for MobileNetV1. The difference is defined to be (EWR - LR) / LR. The data is obtained over 5-10 runs over combinations of different setups: noise level $\sigma$ and $2\sigma$, noisy gradient proportion 10\% and 25\%, $\varepsilon=1,2$. The target sparsity is 0.95.}
    \label{fig:accuracy_mobilenetv1}
\end{figure}
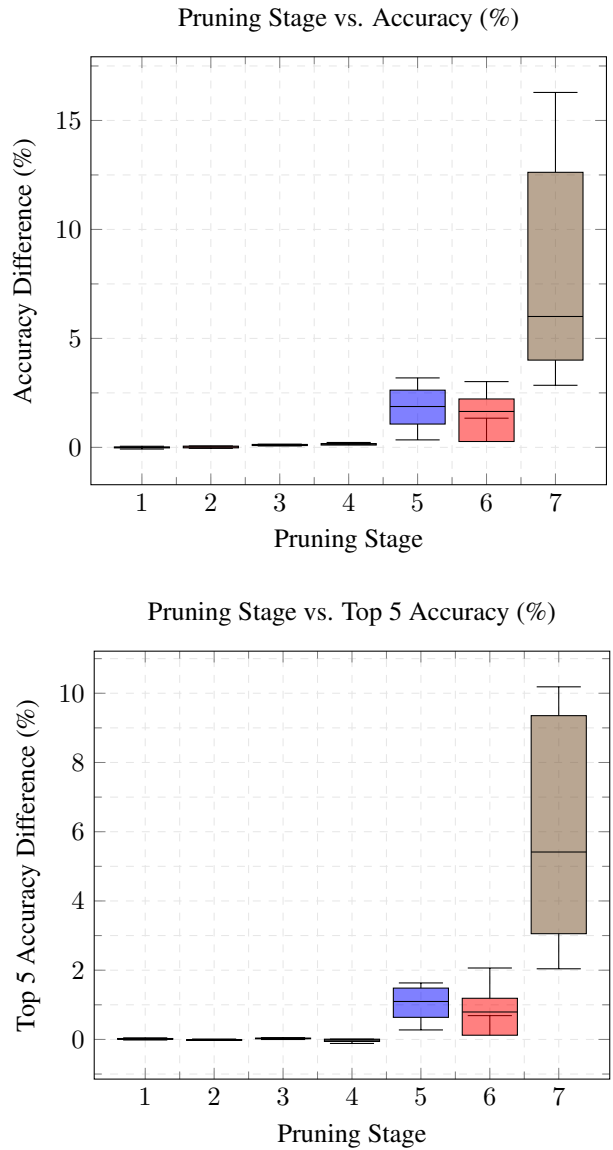

\begin{table}[t]
\centering
\caption{Comparison of testing Top-1 accuracy (\%) (without the fine-tuning steps imposed) values for LR and EWR for ResNet20. The result is averaged over 25 runs. The 90\% confidence interval is reported. The target sparsity is set to be 0.95.}
\label{tab:accuracy_resnet20}
\begin{tabular}{c|*{4}{ccc}}
\toprule
& \multicolumn{6}{c}{10\% Noisy Data} \\
  Sparsity       & \multicolumn{3}{c}{Noise = \(\sigma\)} & \multicolumn{3}{c}{Noise = \(2\sigma\)} \\
\cmidrule(lr){2-4} \cmidrule(lr){5-7}
      & LR   & EWR  & Diff & LR   & EWR  & Diff \\
\midrule
0.95 & 11.65 ($\pm$0.21) & 11.70 ($\pm$0.15)  & 0.47\% & 11.53 ($\pm$0.14)  & \textbf{11.72} ($\pm$0.14) & \textcolor{teal}{\textbf{1.70}\%} \\
0.84 & 57.01 ($\pm$0.10)  & \textbf{58.17} ($\pm$0.10) & \textcolor{teal}{\textbf{2.03}\%} & 56.03 ($\pm$0.09)  & \textbf{57.50} ($\pm$0.09)  & \textcolor{teal}{\textbf{2.63}\%} \\
0.74 & 80.53 ($\pm$0.05) & \textbf{81.12} ($\pm$0.07) & \textcolor{teal}{\textbf{0.73}\%} & 80.23 ($\pm$0.05)  & \textbf{81.06} ($\pm$0.06)  & \textcolor{teal}{\textbf{1.03}\%} \\
0.63 & 89.85 ($\pm$0.06)  & 89.90 ($\pm$0.06)        & 0.00\%                              & 88.95 ($\pm$0.06) &    88.96 ($\pm$0.05)         & 0.85\% \\
\bottomrule
\end{tabular}
\begin{tabular}{c|*{4}{ccc}}
& \multicolumn{6}{c}{25\% Noisy Data} \\
  Sparsity       & \multicolumn{3}{c}{Noise = \(\sigma\)} & \multicolumn{3}{c}{Noise = \(2\sigma\)} \\
\cmidrule(lr){2-4} \cmidrule(lr){5-7}
      & LR   & EWR  & Diff & LR   & EWR & Diff \\
\midrule
0.95 & 11.63 ($\pm$0.17) & 11.98 ($\pm$0.15) & \textcolor{teal}{\textbf{3.03}\%} & 11.68 ($\pm$0.18) & \textbf{12.04} ($\pm$0.16) & \textcolor{teal}{\textbf{3.04}\%} \\
0.84 & 56.90 ($\pm$0.09) & \textbf{58.90} ($\pm$0.09) & \textcolor{teal}{\textbf{3.52}\%} & 57.90 ($\pm$0.07) & \textbf{59.76} ($\pm$0.06) & \textcolor{teal}{\textbf{3.22}\%} \\
0.74 & 80.37 ($\pm$0.05) & \textbf{80.76} ($\pm$0.10) & \textcolor{teal}{\textbf{0.49}\%} & 81.50 ($\pm$0.08) & \textbf{82.27} ($\pm$0.05) & \textcolor{teal}{\textbf{0.95}\%} \\
0.63 & 89.62 ($\pm$0.05) & 89.62 ($\pm$0.09)          &                            0.00\% & 90.56 ($\pm$0.06) &          90.56 ($\pm$0.05) & 0.00\% \\
\bottomrule
\end{tabular}
\end{table}
\begin{table}[!h]
\centering
\caption{Comparison of testing Top-1 accuracy (\%) (without the fine-tuning step imposed) values for LR and EWR for MobileNetV1. The result is averaged from 10 runs. The 90\% confidence interval is reported. The target sparsity is set to be 0.75.}
\label{tab:accuracy_mobilenetv1}
\begin{tabular}{c|*{4}{ccc}}
\toprule
& \multicolumn{6}{c}{10\% Noisy Data} \\
  Sparsity       & \multicolumn{3}{c}{Noise = \(\sigma\)} & \multicolumn{3}{c}{Noise = \(2\sigma\)} \\
\cmidrule(lr){2-4} \cmidrule(lr){5-7}
      & LR   & EWR  & Diff & LR   & EWR  & Diff \\
\midrule
0.75 &   17.10 ($\pm$0.20)   &  \textbf{17.79} ($\pm$0.15)  & \textcolor{teal}{\textbf{4.02}\%}  &   17.05 ($\pm$0.21)  &   \textbf{17.84} ($\pm$0.18)  &  \textcolor{teal}{\textbf{4.62}\%}   \\
0.63 &   43.92 ($\pm$0.05)  &  \textbf{44.92} ($\pm$0.05)   & \textcolor{teal}{\textbf{2.27}\%}  &   43.27 ($\pm$0.10)  &   \textbf{44.53} ($\pm$0.10)  &  \textcolor{teal}{\textbf{2.92}\%}       \\
0.53 &   61.06 ($\pm$0.04)  &  \textbf{61.84} ($\pm$0.05)   & \textcolor{teal}{\textbf{1.28}\%}  &   60.51 ($\pm$0.05)  &   \textbf{63.90} ($\pm$0.05)  &    \textcolor{teal}{\textbf{2.30}\%}       \\
0.42 &   68.09 ($\pm$0.05)  &  68.09 ($\pm$0.06)            &    0.00\%                          &   67.24 ($\pm$0.05)  &   67.24 ($\pm$0.05)  &      0.00\%     \\
\bottomrule
\end{tabular}
\begin{tabular}{c|*{4}{ccc}}
& \multicolumn{6}{c}{25\% Noisy Data} \\
  Sparsity       & \multicolumn{3}{c}{Noise = \(\sigma\)} & \multicolumn{3}{c}{Noise = \(2\sigma\)} \\
\cmidrule(lr){2-4} \cmidrule(lr){5-7}
      & LR   & EWR  & Diff & LR   & EWR  & Diff \\
\midrule
0.75 &   13.97 ($\pm$0.14)   &  \textbf{14.88} ($\pm$0.26)   &  \textcolor{teal}{\textbf{6.52}\%}  &  13.78 ($\pm$0.21)   &  \textbf{14.78} ($\pm$0.25)  &     \textcolor{teal}{\textbf{7.29}\%}       \\
0.63 &   43.45 ($\pm$0.10)   &  \textbf{44.78} ($\pm$0.09)    & \textcolor{teal}{\textbf{3.05}\%}  &  43.21 ($\pm$0.10)   &  \textbf{45.13} ($\pm$0.11)  &    \textcolor{teal}{\textbf{4.44}\%}       \\
0.53 &   60.59 ($\pm$0.05)   &  \textbf{61.26} ($\pm$0.04)    & \textcolor{teal}{\textbf{1.10}\%}  &  60.07 ($\pm$0.05)   &  \textbf{61.56} ($\pm$0.05)  &       \textcolor{teal}{\textbf{2.48}\%}     \\
0.42 &   67.33 ($\pm$0.04)   &  67.46 ($\pm$0.05)             &   0.20\%                           &  66.92 ($\pm$0.06)   &  66.92 ($\pm$0.05)  &    0.00\%       \\
\bottomrule
\end{tabular}
\end{table}

The left plot, representing data with a 10\% noise level, depicts a prominent decrease in the difference of loss as sparsity is reduced for both noise levels \(2\sigma\) and \(\sigma\). Notably, in the \(2\sigma\) noise scenario, there is a sharp decline in loss difference when sparsity transitions from 0.95 to 0.53. Beyond this threshold, the loss difference stabilizes and remains near zero. For the noise level \(\sigma\), the decrease appears more gradual. It is of interest to observe that the loss difference diminishes more swiftly for \(2\sigma\) compared to \(\sigma\). The error bars offer insights into data variability, showcasing broader intervals at elevated sparsity levels, which suggests greater unpredictability at these levels, particularly in the \(2\sigma\) setting.

The right plot represents data with 25\% noise. While the trends in loss difference share similarities with the 10\% noisy data, the exact values differ slightly. In this 25\% noise setting, the decline in loss difference between the two noise levels is similar. The error bars, indicating confidence intervals, highlight the increased variability at larger sparsity levels. This variability is most noticeable for the \(2\sigma\) setting at the top sparsity levels.

\figurename~\ref{fig:loss_diff_mobilenetv1} depicts the difference in loss between LR and EWR algorithms applied to the MobileNetV1. In both two cases 10\% and 25\% noisy data, as sparsity increases, the loss difference diminishes, particularly when sparsity is approximately 0.42 or less. The 10\% noisy data reveals that the difference in loss for noise level $2\sigma$ is marginally higher than that of $\sigma$ for most sparsity levels. In contrast, the 25\% noisy data sometimes exhibits a reversal in this trend, especially at the highest sparsity level of 0.74.

Confidence intervals provided at select data points underscore the reliability of the data, with the 25\% noisy data showing tighter intervals compared to the 10\% scenario. This infers a higher consistency in the measurements or a minimized effect of outliers in the 25\% noisy data. These plots accentuate the interplay between sparsity, noise, and the performance difference between the two algorithms, emphasizing the significance of noise levels in algorithmic performance evaluations across various sparsity conditions. The underlying rationale is that as the noise intensity is too high, the performance of both LR and EWR tends to deteriorate. Consequently, their performances converge, resulting in a diminished differential between the two.

\textbf{Analysis of Accuracy}. The box plots in Figure \ref{fig:accuracy_resnet20} together with Table \ref{fig:accuracy_resnet20} show how the ResNet20 model performs at different pruning stages in terms of top-1 accuracy (also referred to as accuracy or the overall accuracy) and top-5 accuracy difference. Top-5 accuracy means any of our model’s top 5 highest probability answers match with the expected answer. The difference is computed for EWR, using LR as the baseline. When we look closely, we can see patterns that help us understand how much pruning affects the model.

In the earlier pruning stages, both the overall accuracy and the top 5 accuracy differences for EWR are minimal, suggesting that EWR remains closely aligned with the baseline LR in terms of performance. This minimal deviation can be viewed as an advantage, as it implies that even with simplifications brought by pruning, EWR retains its effectiveness compared to LR.

However, as pruning intensifies, the patterns begin to reveal more about EWR's relative strengths. Although the accuracy difference increases, this increase in the context of the baseline suggests that EWR might be better at handling intense pruning than LR. Notably, in the top 5 accuracy, the discrepancies remain relatively low compared to the overall accuracy up until the more aggressive pruning stages. This suggests that while the model's primary prediction confidence may decrease, the true class is still frequently among its top 5 predictions. In essence, EWR seems to retain a broader spectrum of potential correct classifications even when it's uncertain about the primary prediction.

\figurename~\ref{fig:accuracy_mobilenetv1} together with Table \ref{fig:accuracy_mobilenetv1} shows the performance difference between EWR and LR using MobileNetV1 across various pruning stages. In the first plot showcasing the overall accuracy, there's an evident upward trend in the median accuracy difference as pruning intensifies. Initially, the difference is marginal, which suggests that EWR's performance closely mirrors LR during the early pruning stages. However, as we progress to the 5th and 6th stages, the accuracy difference widens considerably, indicating that EWR might be outpacing LR. The 7th stage is particularly striking, with a median accuracy difference surpassing 6\%, pointing towards a potential superiority of EWR in extreme pruning scenarios. 

The second plot focuses on the top 5 accuracy differences, presenting a more varied pattern. The early stages indicate a tight performance race between EWR and LR. By the 2nd stage, there's a minor dip, hinting at EWR's possible underperformance. This scenario changes by the 3rd stage as EWR regains momentum. The later stages, especially the 5th and 6th, denote a significant rise in the median difference in EWR's favor. Much like the accuracy chart, the 7th stage is distinct, with EWR showcasing a considerable advantage in the top 5 accuracy over LR.


\subsection{Algorithm Scalability}
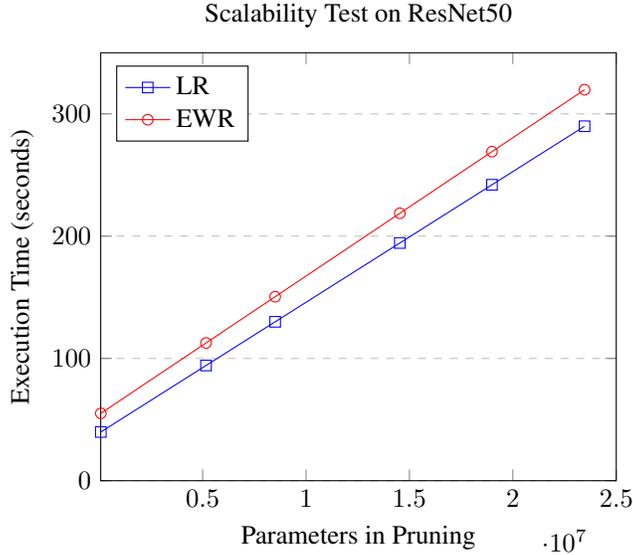
\begin{figure}
    \centering
\begin{tikzpicture}
\begin{axis}[
    title={Scalability Test on ResNet50},
    xlabel=Parameters in Pruning,
    ylabel={Execution Time (seconds)},
    xmin=70000, xmax=25000000,
    ymin=0, ymax=350,
    legend pos=north west,
    legend cell align={left},
    ymajorgrids=true,
    grid style=dashed,
]

\addplot[
    color=blue,
    mark=square,
    ]
    coordinates {
    (75456, 39.85)(5162688, 94.23)(8505024, 129.957922705314)(14534336, 194.4082930756844)(18990784, 242.04552334943642)(23467712, 289.9016747181965)
    };
    \addlegendentry{LR}

\addplot[
    color=red,
    mark=o,
    ]
    coordinates {
(75456, 55.083491847826096)(5162688, 112.65349184782609)(8505024, 150.47725996376812)(14534336, 218.70837107487924)(18990784, 269.14006189613525)(23467712, 319.80351600241545)
    };
    \addlegendentry{EWR}

\end{axis}
\end{tikzpicture}
    \caption{Scalability of Algorithm \ref{alg:cap}. This plot shows the comparison between LR and EWR of the single-round execution time of Lines \ref{alg:cap-G}--\ref{alg:cap-bw}. The test is performed on an NVIDIA Tesla V100 32GB GPU. The fisher sample size is set to 1000. Layers of the networks are gradually added for pruning (hence the number of model parameters $p$ increases). The \gls{ot} planning $\vec{\Pi}$ is solved using the Sinkorn-Knopp method shown in Algorithm \ref{alg:sinkhorn}.}
    \label{fig:scalability}
\end{figure}

In this section, we analyze the scalability of Algorithm \ref{alg:cap} with respect to the number of model parameters involved in pruning. The result is shown in \figurename~\ref{fig:scalability}. It can be observed that the execution time scales linearly with the number of pruning parameters. The extra cost of solving the \gls{ot} is marginal. Theoretically, one could derive this linear scalability by inspecting Line \ref{alg:cap-sgd}, which is the most time-consuming step. The required operations can be decomposed as sequential operations: matrix-vector multiplications $\vec{Gw}$ and $\vec{G\bw}$ in $O(np)$, the vector subtraction $\vec{Gw}-\vec{G\bw}$ in $O(n)$, a matrix-matrix multiplication $\vec{\Pi}(\vec{Gw}-\vec{G\bw})$ in $O(n^2)$, a matrix transposition and multiplication with $\vec{G}$ in $O(np)$, and the vector subtraction and scalar multiplication $\lambda(\vec{w}-\vec{\bw})$ in $O(p)$. Thus, the overall complexity is $O(np)$, with $p$ significantly larger than $n$ practically. Given fixed fisher sample size $n$, the loop of Algorithm \ref{alg:cap} scales linearly with the number of pruning parameters $p$.

\subsection{Optimal Transport Visualization}
\label{sec:ot_vis}

\begin{figure}[!h]
  \centering
\begin{minipage}{0.49\textwidth}
  \begin{tikzpicture}
    \begin{axis}[
    width=\textwidth,
      ybar,
      ymin=0,
      ylabel =Frequency,
      xtick align=inside,
    ]
      \addplot+[
        hist={
          bins=20,
          data min=-50,
          data max=50,
        },
        fill opacity = 0.5
      ] table [y index=0] {
        data
-44.7287654876709
-40.51976413726807
-36.99977560043335
-33.32237586975098
-31.19666452407837
-26.76617612838745
-24.343922996520995
-21.572545433044432
-19.602397918701172
-17.836676788330077
-16.26714038848877
-14.674180459976196
-13.406735491752624
-11.845200657844543
-10.886565256118775
-9.99955141544342
-9.064515542984008
-8.359427118301392
-7.507229137420654
-6.717669749259949
-6.014764034748078
-5.717140853404999
-4.981425034999847
-4.553458750247955
-3.7969252824783326
-3.268317449092865
-2.7421424806118013
-2.4148666679859163
-2.2082783043384553
-2.0401006817817686
-1.832761687040329
-1.6901933073997497
-1.5799229264259338
-1.4075769066810608
-1.2710857808589935
-1.139334389567375
-1.0514077365398407
-0.9507339924573899
-0.866395303606987
-0.7843049809336662
-0.7462189465761184
-0.6721372023224831
-0.5939740851521492
-0.5168060101568699
-0.4764407366514206
-0.42014485634863374
-0.36287571378052236
-0.3122792258858681
-0.2795020941644907
-0.2351204317063093
-0.20391342658549547
-0.17127396762371064
-0.1410866232588887
-0.1197930896654725
-0.10032101469114423
-0.08311589313670993
-0.07390007525682449
-0.06220915745943785
-0.0514072023332119
-0.04489886187948287
-0.036743061942979695
-0.03082720171660185
-0.026503977179527283
-0.02327309069223702
-0.018230963440146297
-0.014246487821219489
-0.010964704595971852
-0.005000715423375368
0.0020724925911054016
0.004769580040010624
0.02707901221292559
0.03956395607383456
0.12323609062150354
0.19780277348363598
0.2663107167776616
0.4310261692095082
0.5579282603561296
0.8132332317322835
1.2316503081894439
1.4686162407232903
2.4280758896515424
3.9381820226528363
6.160088203571569
8.189679229726876
11.398991084098816
13.25673608481884
17.16111854016781
24.517021968960762
31.063429987430574
41.53503165245056
49.650986957550046

      };
\addlegendentry{$\{y_i\}_{i=1}^n$}

      \addplot+[
        hist={
          bins=20,
          data min=-50,
          data max=50,
        },
        fill opacity = 0.5
      ] table [y index=0] {
        data
-26.446129989624023
-14.666034317016601
-11.027981567382813
-9.966349792480468
-7.178849601745606
-6.32210636138916
-5.483050060272217
-4.589105796813965
-4.252249479293823
-3.455105924606323
-2.9358728647232057
-2.54457049369812
-2.1918737173080443
-2.0607580661773683
-2.0019356489181517
-1.811099934577942
-1.6709949254989624
-1.5510800004005432
-1.3855066061019898
-1.1773338675498963
-1.034887146949768
-0.9163250803947449
-0.8503084063529969
-0.7939511656761169
-0.6657961547374726
-0.5949825286865235
-0.5366455733776092
-0.4677053153514862
-0.40906850099563596
-0.38017277121543885
-0.35293337106704714
-0.3062705993652344
-0.26038987934589386
-0.23552284240722657
-0.2087348848581314
-0.19074958562850952
-0.16419025659561157
-0.13606248646974564
-0.12898206412792207
-0.12007313966751099
-0.10304968282580376
-0.09152087047696114
-0.08348303586244583
-0.07347048074007034
-0.06379484049975873
-0.05966637171804905
-0.050616918690502645
-0.04299906734377146
-0.036661638505756856
-0.030570358224213123
-0.023534950241446494
-0.019247822090983392
-0.017457037791609765
-0.014926230907440186
-0.012526667583733797
-0.011260464508086442
-0.00864205751568079
-0.007053696550428867
-0.005188631452620029
-0.0050029810285195705
-0.004552600486204028
-0.003634898574091494
-0.002780522033572197
-0.0023107350221835076
-0.001970779651310295
-0.0013321380014531315
-0.001051419647410512
-0.0008626233611721545
-0.0007267628272529692
-0.00037601017393171785
0.00111329790961463
0.005663557507796213
0.013124405026610475
0.016015733753010863
0.017169956707948585
0.037158070382429284
0.054782909000641665
0.07290912889257015
0.08367828428235953
0.34764486899184704
0.5373325018117612
0.6434316824293432
0.8201657394576106
1.2662785430525019
1.419892506301403
1.5362198233604432
1.9270047903060914
2.3871057510375975
2.6115083515644075
3.213484501838684
3.582184338569641
4.328614664077759
5.515989065170288
7.291481637954712
9.070875930786134
10.452131414413453
12.699311542510987
18.308994674682616
28.869419860839844
37.0594970703125
      };
\addlegendentry{$\{x_i\}_{i=1}^n$}
\end{axis}
  \end{tikzpicture}
\end{minipage}
\begin{minipage}{0.49\textwidth}
\begin{tikzpicture}
    \begin{axis}[
    width=\textwidth,
      ymin=0,
      ylabel=Density,
      ylabel near ticks, yticklabel pos=right,
    yticklabel style={
        /pgf/number format/precision=2,
        /pgf/number format/fixed},
    ]
      \addplot[smooth, blue, very thick, opacity=0.5] coordinates {
( -44.7287654876709 , 0.003989990561964545 )
( -40.51976413726807 , 0.003998125884443457 )
( -36.99977560043335 , 0.004002175641771618 )
( -33.32237586975098 , 0.004410619474054185 )
( -31.19666452407837 , 0.004406220574183096 )
( -26.76617612838745 , 0.0042018925153823996 )
( -24.343922996520995 , 0.0042874455125890484 )
( -21.572545433044432 , 0.004651744471610059 )
( -19.602397918701172 , 0.005416948368752738 )
( -17.836676788330077 , 0.006023549359040982 )
( -16.26714038848877 , 0.0063575120132883895 )
( -14.674180459976196 , 0.0070009595600526425 )
( -13.406735491752624 , 0.007200911347274012 )
( -11.845200657844543 , 0.00858059943048402 )
( -10.886565256118775 , 0.010307003792963898 )
( -9.99955141544342 , 0.011234778879167216 )
( -9.064515542984008 , 0.012014544927056121 )
( -8.359427118301392 , 0.012517598135099466 )
( -7.507229137420654 , 0.013397249094338383 )
( -6.717669749259949 , 0.015090571003733637 )
( -6.014764034748078 , 0.016701196835400077 )
( -5.717140853404999 , 0.017148540859321534 )
( -4.981425034999847 , 0.017661698190393725 )
( -4.553458750247955 , 0.018121227473611112 )
( -3.7969252824783326 , 0.021656087686393006 )
( -3.268317449092865 , 0.02843236477535534 )
( -2.7421424806118013 , 0.04109010191088578 )
( -2.4148666679859163 , 0.053092656662065606 )
( -2.2082783043384553 , 0.06265708440459075 )
( -2.0401006817817686 , 0.07166341822441544 )
( -1.832761687040329 , 0.08425413985958947 )
( -1.6901933073997497 , 0.09377516525101587 )
( -1.5799229264259338 , 0.10153416733328603 )
( -1.4075769066810608 , 0.1141275873651857 )
( -1.2710857808589935 , 0.12424925173636174 )
( -1.139334389567375 , 0.1338650958750073 )
( -1.0514077365398407 , 0.14005568923252354 )
( -0.9507339924573899 , 0.14678440496137898 )
( -0.866395303606987 , 0.15202682675034263 )
( -0.7843049809336662 , 0.15669875536248123 )
( -0.7462189465761184 , 0.15870052366843998 )
( -0.6721372023224831 , 0.1622575507644729 )
( -0.5939740851521492 , 0.16548130085251087 )
( -0.5168060101568699 , 0.16807826024993552 )
( -0.4764407366514206 , 0.16918895896323652 )
( -0.42014485634863374 , 0.17043979240279486 )
( -0.36287571378052236 , 0.17134323576010763 )
( -0.3122792258858681 , 0.17182247980608678 )
( -0.2795020941644907 , 0.17197013451297802 )
( -0.2351204317063093 , 0.17196348881443815 )
( -0.20391342658549547 , 0.17181552626004756 )
( -0.17127396762371064 , 0.1715337160151678 )
( -0.1410866232588887 , 0.17115738979945083 )
( -0.1197930896654725 , 0.170825180312226 )
( -0.10032101469114423 , 0.17047319120199936 )
( -0.08311589313670993 , 0.17012400087079188 )
( -0.07390007525682449 , 0.16992227801722906 )
( -0.06220915745943785 , 0.16965169048541143 )
( -0.0514072023332119 , 0.16938712321069735 )
( -0.04489886187948287 , 0.16922098947097136 )
( -0.036743061942979695 , 0.16900568099173385 )
( -0.03082720171660185 , 0.16884456141585832 )
( -0.026503977179527283 , 0.16872419468567373 )
( -0.02327309069223702 , 0.168632797139794 )
( -0.018230963440146297 , 0.16848769899746124 )
( -0.014246487821219489 , 0.16837091712287833 )
( -0.010964704595971852 , 0.16827332754916272 )
( -0.005000715423375368 , 0.16809273984175224 )
( 0.0020724925911054016 , 0.16787316382471967 )
( 0.004769580040010624 , 0.16778789750646522 )
( 0.02707901221292559 , 0.16705017887613693 )
( 0.03956395607383456 , 0.16661225205987218 )
( 0.12323609062150354 , 0.16322398403501653 )
( 0.19780277348363598 , 0.15956765058440306 )
( 0.2663107167776616 , 0.15571613849585544 )
( 0.4310261692095082 , 0.14476534621907725 )
( 0.5579282603561296 , 0.13501123750548974 )
( 0.8132332317322835 , 0.11329349169690203 )
( 1.2316503081894439 , 0.07732116435885203 )
( 1.4686162407232903 , 0.05953480175034487 )
( 2.4280758896515424 , 0.017896557713335266 )
( 3.9381820226528363 , 0.005992665761094049 )
( 6.160088203571569 , 0.004839924978769529 )
( 8.189679229726876 , 0.004521708864838392 )
( 11.398991084098816 , 0.004722942798639229 )
( 13.25673608481884 , 0.004701761706242587 )
( 17.16111854016781 , 0.003991375921272462 )
( 24.517021968960762 , 0.003989422805993689 )
( 31.063429987430574 , 0.003989422805986592 )
( 41.53503165245056 , 0.003989422804014345 )
( 49.650986957550046 , 0.003989422804014345 )

      };
\addlegendentry{KDE $y$}

      \addplot[smooth, red, very thick, opacity=0.5] coordinates {
( -26.446129989624023 , 0.003989422804014327 )
( -14.666034317016601 , 0.0039948183916753 )
( -11.027981567382813 , 0.006267998944835533 )
( -9.966349792480468 , 0.006347603997142836 )
( -7.178849601745606 , 0.007984011777445264 )
( -6.32210636138916 , 0.011004786658521047 )
( -5.483050060272217 , 0.013057190774002938 )
( -4.589105796813965 , 0.01584716172220599 )
( -4.252249479293823 , 0.017483186905167255 )
( -3.455105924606323 , 0.025231209280453786 )
( -2.9358728647232057 , 0.03637972610998607 )
( -2.54457049369812 , 0.05068147918059315 )
( -2.1918737173080443 , 0.07020832797125233 )
( -2.0607580661773683 , 0.07949409143583262 )
( -2.0019356489181517 , 0.08405366939033186 )
( -1.811099934577942 , 0.1005730230104795 )
( -1.6709949254989624 , 0.11435819374904323 )
( -1.5510800004005432 , 0.12718389041893763 )
( -1.3855066061019898 , 0.14618228952912204 )
( -1.1773338675498963 , 0.17136787252540492 )
( -1.034887146949768 , 0.1886773392349693 )
( -0.9163250803947449 , 0.20260536844026072 )
( -0.8503084063529969 , 0.2100114317253472 )
( -0.7939511656761169 , 0.2160667503381168 )
( -0.6657961547374726 , 0.22867288827844715 )
( -0.5949825286865235 , 0.23479901101868814 )
( -0.5366455733776092 , 0.23932216691507646 )
( -0.4677053153514862 , 0.24399642419737866 )
( -0.40906850099563596 , 0.2473557384722815 )
( -0.38017277121543885 , 0.24879149416038307 )
( -0.35293337106704714 , 0.2500076853238894 )
( -0.3062705993652344 , 0.25177358873131456 )
( -0.26038987934589386 , 0.2531090664252198 )
( -0.23552284240722657 , 0.2536634002980366 )
( -0.2087348848581314 , 0.25412543994754766 )
( -0.19074958562850952 , 0.25435641117693447 )
( -0.16419025659561157 , 0.25458033808466574 )
( -0.13606248646974564 , 0.25466444333392835 )
( -0.12898206412792207 , 0.2546607402130939 )
( -0.12007313966751099 , 0.2546418481907043 )
( -0.10304968282580376 , 0.25456161738877375 )
( -0.09152087047696114 , 0.25447436912304505 )
( -0.08348303586244583 , 0.25439781738335215 )
( -0.07347048074007034 , 0.2542843993177259 )
( -0.06379484049975873 , 0.25415577579967213 )
( -0.05966637171804905 , 0.25409520751837 )
( -0.050616918690502645 , 0.25395055614627693 )
( -0.04299906734377146 , 0.25381614358310767 )
( -0.036661638505756856 , 0.25369552648017163 )
( -0.030570358224213123 , 0.2535720730413037 )
( -0.023534950241446494 , 0.2534203218932 )
( -0.019247822090983392 , 0.2533230400812213 )
( -0.017457037791609765 , 0.25328132686033256 )
( -0.014926230907440186 , 0.25322129415641437 )
( -0.012526667583733797 , 0.2531632049679247 )
( -0.011260464508086442 , 0.2531320937633763 )
( -0.00864205751568079 , 0.25306675371691123 )
( -0.007053696550428867 , 0.25302645797168705 )
( -0.005188631452620029 , 0.25297850696015706 )
( -0.0050029810285195705 , 0.2529736963204327 )
( -0.004552600486204028 , 0.252961997654372 )
( -0.003634898574091494 , 0.25293803650557084 )
( -0.002780522033572197 , 0.2529155795445923 )
( -0.0023107350221835076 , 0.2529031700688268 )
( -0.001970779651310295 , 0.25289416298684364 )
( -0.0013321380014531315 , 0.252877180672359 )
( -0.001051419647410512 , 0.2528696905827404 )
( -0.0008626233611721545 , 0.2528646444146072 )
( -0.0007267628272529692 , 0.25286100877498735 )
( -0.00037601017393171785 , 0.2528516057945968 )
( 0.00111329790961463 , 0.2528114105786429 )
( 0.005663557507796213 , 0.252685898558399 )
( 0.013124405026610475 , 0.2524712966641454 )
( 0.016015733753010863 , 0.2523851947343374 )
( 0.017169956707948585 , 0.2523503651858361 )
( 0.037158070382429284 , 0.25170591981583973 )
( 0.054782909000641665 , 0.25107322213788463 )
( 0.07290912889257015 , 0.25036002427247933 )
( 0.08367828428235953 , 0.24990650734293443 )
( 0.34764486899184704 , 0.2323085744854955 )
( 0.5373325018117612 , 0.2131859537542833 )
( 0.6434316824293432 , 0.20079971337790387 )
( 0.8201657394576106 , 0.17851046801070272 )
( 1.2662785430525019 , 0.12149520569486344 )
( 1.419892506301403 , 0.10401328933567089 )
( 1.5362198233604432 , 0.09199864743368533 )
( 1.9270047903060914 , 0.06023189077620182 )
( 2.3871057510375975 , 0.03818969584887133 )
( 2.6115083515644075 , 0.03178131317531083 )
( 3.213484501838684 , 0.021983475281157256 )
( 3.582184338569641 , 0.018448059932334764 )
( 4.328614664077759 , 0.013121853928328613 )
( 5.515989065170288 , 0.007787444541096337 )
( 7.291481637954712 , 0.005715149172366125 )
( 9.070875930786134 , 0.006358165409111461 )
( 10.452131414413453 , 0.005872694614942594 )
( 12.699311542510987 , 0.004314361998249195 )
( 18.308994674682616 , 0.003989423389616247 )
( 28.869419860839844 , 0.0039894228040143346 )
( 37.0594970703125 , 0.0039894228040143346 )
      };
\addlegendentry{KDE $x$}
    \end{axis}
\end{tikzpicture}
\end{minipage}

  \caption{The histogram and kde plot of the empirical distribution $\{x_i\}_{i=1}^n$ and $\{y_i\}_{i=1}^n$.}
  \label{fig:histogram}
\end{figure}
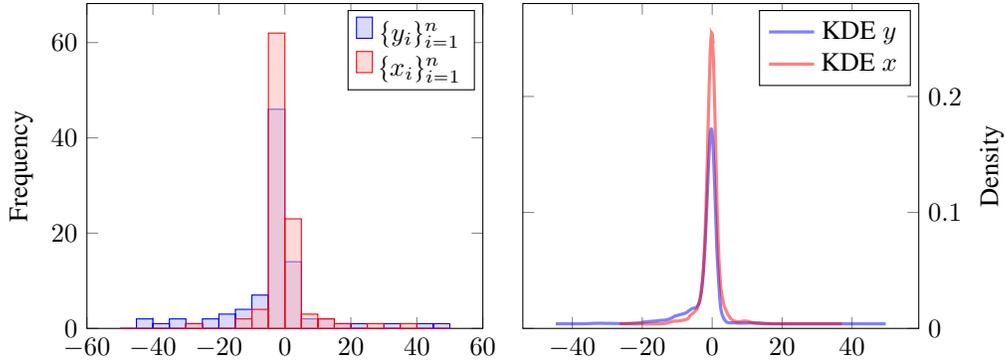

In this section, we showcase the optimized \gls{ot} plan denoted as $\vec{\Pi}$. This was derived from the pruning applied to ResNet20. Figure \ref{fig:histogram} displays two data sets: 1) $\{x_i\}_{i=1}^n$: This is $\vec{G}\vec{w}$ where $\vec{w}$ is the pruned model. 2) $\{y_i\}_{i=1}^n$: This represents $\vec{G}\vec{\bw}$ where $\vec{\bw}$ is the original unpruned model.

\begin{figure}[!h]
    \centering
    \begin{subfigure}{0.45\textwidth}
        \centering
        \includegraphics[width=\textwidth]{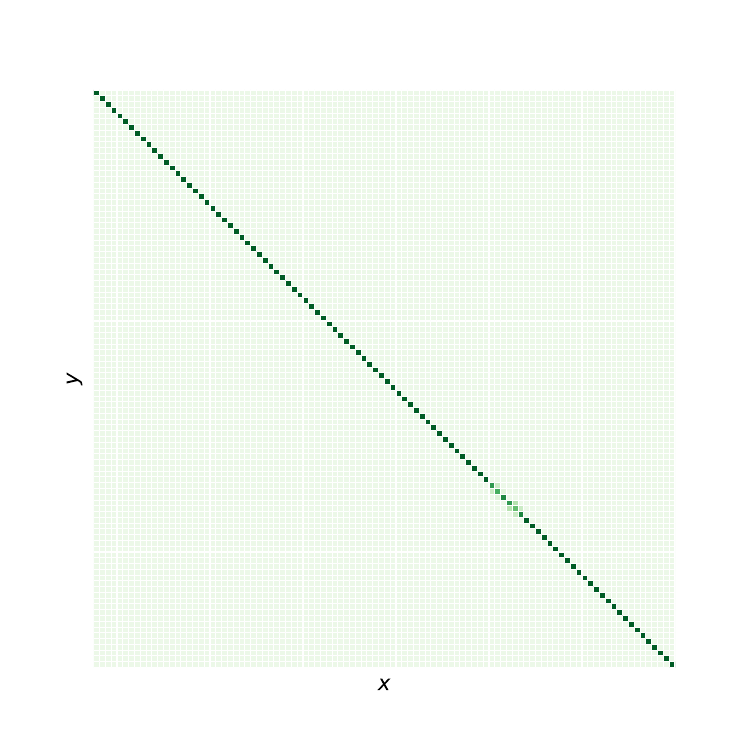}
        \caption{$\varepsilon=0$}
        \label{fig:sub_a}
    \end{subfigure}
    \hfill
    \begin{subfigure}{0.45\textwidth}
        \centering
        \includegraphics[width=\textwidth]{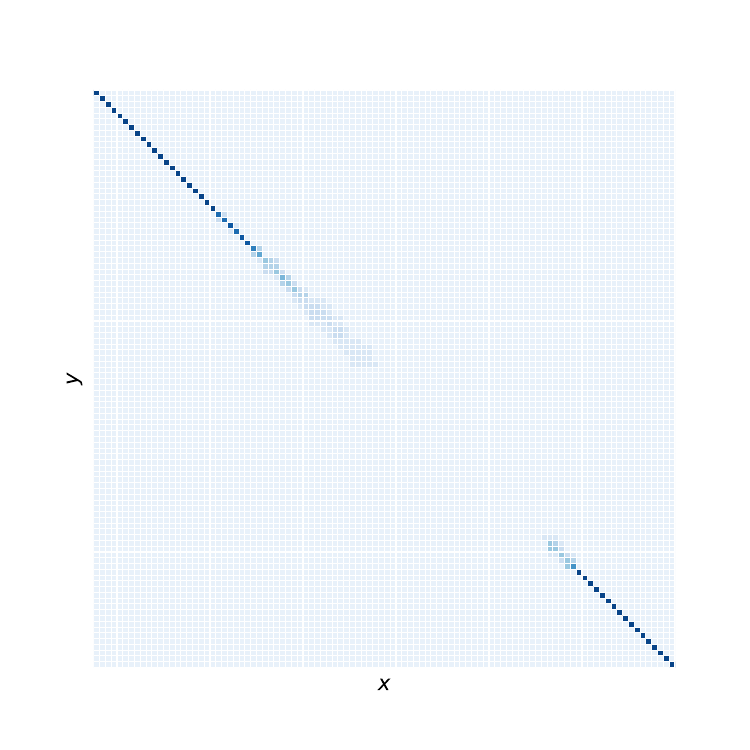}
        \caption{$\varepsilon=0.01$}
        \label{fig:sub_b}
    \end{subfigure}
    
    \begin{subfigure}{0.45\textwidth}
        \centering
        \includegraphics[width=\textwidth]{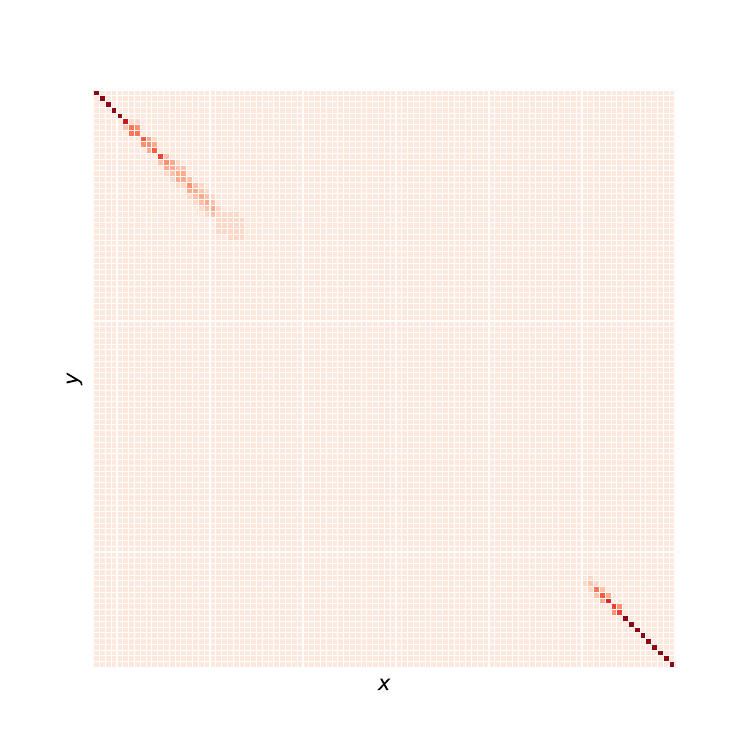}
        \caption{$\varepsilon=1$}
        \label{fig:sub_c}
    \end{subfigure}
    \hfill
    \begin{subfigure}{0.45\textwidth}   
        \centering
        \includegraphics[width=\textwidth]{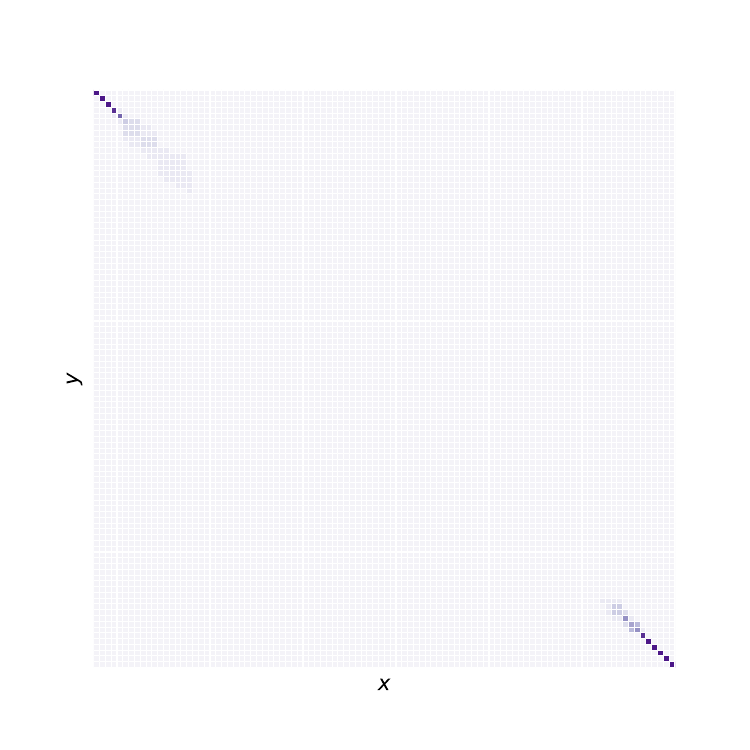}
        \caption{$\varepsilon=10$}
        \label{fig:sub_d}
    \end{subfigure}
    
    \caption{Heatmap of the optimized \gls{ot} plan $\vec{\Pi}$ across different $\varepsilon$. Darker color indicate larger value in $\vec{\Pi}$.}
    \label{fig:ot-matrix}
\end{figure}

In Figure \ref{fig:ot-matrix}, the matrix $\vec{\Pi}$ is shown through vibrant heatmaps that adjust with varying $\varepsilon$ values.
For small $\varepsilon$, the majority of the data remains near a diagonal. As $\varepsilon$ increases, there's a broader data distribution, notably for central data points.

\end{document}